\documentclass[twocolumn,english]{IEEEtran}
\usepackage[T1]{fontenc}
\usepackage[utf8]{inputenc}
\usepackage{fancyhdr}
\usepackage{color}
\usepackage{babel}
\usepackage{array}
\usepackage{float}
\usepackage{multirow}
\usepackage{varwidth}
\usepackage{amstext}
\usepackage{amssymb}
\usepackage{graphicx}
\usepackage[bookmarks=true,bookmarksnumbered=false,bookmarksopen=false,
 breaklinks=false,pdfborder={0 0 1},backref=false,colorlinks=true]
 {hyperref}
\hypersetup{pdftitle={NDELS: A Novel Approach for Nighttime Dehazing, Low-Light Enhancement, and Light Suppression},
 pdfauthor={Silvano A. Bernabel and Sos S. Agaian},
 pdfsubject={Nighttime Dehazing},
 pdfkeywords={Single-image nighttime dehazing, non-uniform haze, bright light suppression, low-light enhancement, multiscale retinex}}

\makeatletter

\providecommand{\tabularnewline}{\\}
\newenvironment{cellvarwidth}[1][t]
    {\begin{varwidth}[#1]{\linewidth}}
    {\@finalstrut\@arstrutbox\end{varwidth}}

\newcommand{\lyxaddress}[1]{
	\par {\raggedright #1
	\vspace{1.4em}
	\noindent\par}
}

\pdfoutput=1
\usepackage{cite}
\ifdefined\showcaptionsetup
 \PassOptionsToPackage{caption=false}{subfig}
\fi
\usepackage{subfig}
\makeatother

\begin{document}
\title{NDELS: A Novel Approach for Nighttime Dehazing, Low-Light Enhancement,
and Light Suppression}
\author{Silvano A. Bernabel and Sos S. Agaian (Fellow, IEEE){\small}\\
{\small The Graduate Center, City University of New York, New York,
NY USA}}
\maketitle
\begin{abstract}
This paper tackles the intricate challenge of improving the quality
of nighttime images under hazy and low-light conditions. Overcoming
issues including nonuniform illumination glows, texture blurring,
glow effects, color distortion, noise disturbance, and overall, low
light have proven daunting. Despite the inherent difficulties, this
paper introduces a pioneering solution named Nighttime Dehazing, Low-Light
Enhancement, and Light Suppression (NDELS). NDELS utilizes a unique
network that combines three essential processes to enhance visibility,
brighten low-light regions, and effectively suppress glare from bright
light sources. In contrast to limited progress in nighttime dehazing,
unlike its daytime counterpart, NDELS presents a comprehensive and
innovative approach. The efficacy of NDELS is rigorously validated
through extensive comparisons with eight state-of-the-art algorithms
across four diverse datasets. Experimental results showcase the superior
performance of our method, demonstrating its outperformance in terms
of overall image quality, including color and edge enhancement. Quantitative
(PSNR, SSIM) and qualitative metrics (CLIPIQA, MANIQA, TRES), measure
these results.
\end{abstract}

\begin{IEEEkeywords}
Single-image nighttime dehazing, non-uniform haze, bright light suppression,
low-light enhancement, multiscale retinex
\end{IEEEkeywords}

\lyxaddress{This work was supported in part by the U.S. Department of Transportation,
Federal Highway Administration (FHWA), under Contract 693JJ320C000023}

\thispagestyle{empty}

\section*{Introduction}

Nighttime dehazing is a crucial and dynamic research field that addresses
a pressing issue in low-light image capture. Poor quality images captured
under low-light conditions can negatively impact applications, ascribable
to non-uniform illumination, texture blurring, glow effects, color
distortion, haze density, noise disturbance, and other light sources.
These degradations render images useless, reducing user experience
and limiting the extent to which applications may benefit from higher
image quality.

By improving the quality through dehazing techniques, images can be
effectively utilized in computer vision, surveillance, and autonomous
systems \cite{saravanarajan_improving_2023,guo_heterogeneous_2021}.
Despite atmospheric conditions, including haze and foggy weather,
Qiu et al. \cite{qiu_idod-yolov7_2023}, demonstrated that effect
suppression improves autonomous driving. In addition, Liu et al. \cite{liu_analysis_2020},
showed advantages of decreasing haze density, one of which is increasing
object detection accuracy. 

Despite having achieved satisfying dehazing results in daytime settings,
in nighttime, daytime methods have been delivering poor quality, at
least due to differences in degradation characteristics. Nevertheless,
there have been developments. One strategy, first transformed color
characteristics between night and day domains, then used prior-based
dehazing methods \cite{pei_nighttime_2012}. Another, had first reduced
the glowing effects by estimating and removing a glow layer, then
applied dehazing \cite{trongtirakul_transmission_2022}. The remaining
daytime dehazing algorithms are either prior-based \cite{kaiming_he_single_2011,fattal_dehazing_2014,zhu_fast_2015,bui_single_2018,ju_idgcp_2019,ju_idrlp_2021,trongtirakul_single_2020,berman_single_2020}
or learning-based \cite{cai_dehazenet_2016,ren_single_2020,li_aod-net_2017,ren_gated_2018,yang_proximal_2018,li_single_2018,qu_enhanced_2019,ren_single_2016,dong_multi-scale_2020,qin_ffa-net_2020,zhang_drcdn_2020,zhang_joint_2020,zhang_nldn_2020,shao_domain_2020,chen_psd_2021,wu_contrastive_2021,zhang_aidednet_2023,trongtirakul_transmission_2022}.
These methods have shown good daytime dehazing. But, when faced with
nighttime dehazing, quality still eludes them.

Nighttime-dehazing methods are naturally better, but not perfect.
Recently, Zhang et al. \cite{zhang_nighttime_2014,zhang_fast_2017},
developed maximum reflection-prior and parameter estimation to achieve
dehazing. Similarly, Wang et al., designed gray haze line prior (GHLP)
\cite{wang_variational_2022}, and a variation method to attain dehazing.
Later, Zhang et al. \cite{zhang_nighttime_2020}, created an optimal-scale-fusion
method and a novel, benchmark dataset of nighttime images with synthetic
haze. These methods improve dehazing, increase visibility and reduce
light glow. Despite this, they amplify noise and, as a result, reduce
the visual quality.

This paper addresses the challenge of enhancing the quality of hazy-nighttime
images in low light. Quality that has been degraded by non-uniform
illumination, texture blurring, glow effects, color distortion, noise
disturbance, and low light. The paper proposes a novel nighttime dehazing,
enhancement, and light-suppression network (NDELS) to engage the problem.
This contribution provides the following:
\begin{enumerate}
\item Introduction of an end-to-end single-image, nighttime-dehazing, convolutional
neural network, providing a holistic solution to the challenges posed
by nighttime conditions.
\item We propose a novel approach for generating training data, crucial
for effectively suppressing light effects, and enhancing the adaptability
and robustness of the NDELS architecture.
\item Conducting detailed computer simulations to validate the effectiveness
of the NDELS architecture, ensuring its practical applicability and
superior performance in real-world scenarios.
\end{enumerate}
The remainder of this paper is organized as follows. The background
section presents existing dehazing studies with their limitations
and advantages. Second, the proposed-method section describes the
NDELS architecture. Third, the experimental results, including quantitative
and qualitative comparisons with state-of-the-art, nighttime-dehazing
and enhancement methods, an ablation study, a real-world application,
a subjective study, runtime estimates, and limitations. Finally, we
conclude the paper with final remarks.

\begin{figure*}
\centering
\subfloat[Hazy image]{\includegraphics[scale=0.2]{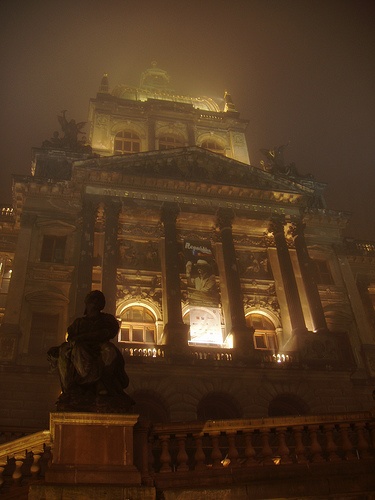}

}\subfloat[Low-light Module]{\includegraphics[scale=0.2]{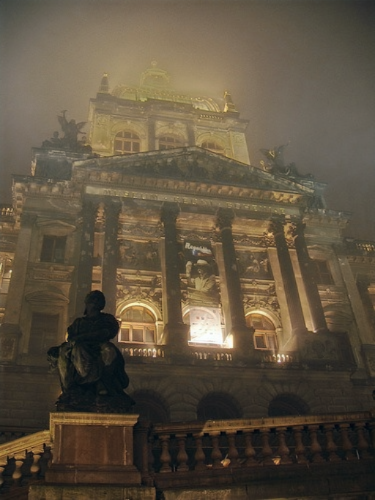}

}\subfloat[Dehazing Module]{\includegraphics[scale=0.2]{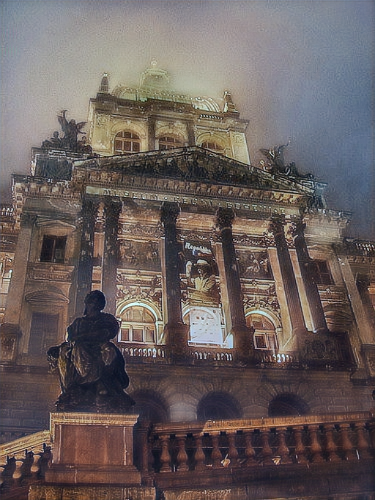}

}\subfloat[Multiscale Retinex]{\includegraphics[scale=0.2]{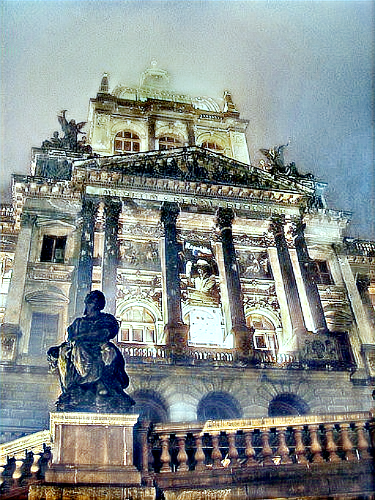}

}\subfloat[Enhancement]{\includegraphics[scale=0.2]{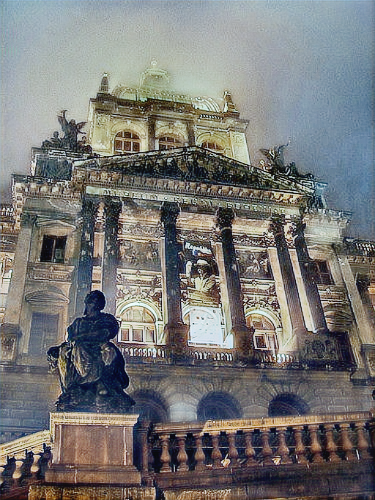}

}

\caption{Breakdown of NDELS network into base components.\protect\label{fig:Network-base-components}}
\end{figure*}

\section*{Background}

This section briefly reviews existing daytime and nighttime-dehazing
methods.

\subsection*{Methods Base on a Physical Model}

These methods have relied primarily on the atmospheric scattering
model (ASM) \cite{koschmieder_theorie_1924}, tuning prior-based estimation
parameters to approximate transmission and atmospheric light.

\subsubsection*{Daytime}

By estimation and optimization, this kind of dehazing has yielded
good results. He et al. \cite{kaiming_he_single_2011}, developed
a dark channel prior (DCP) based on statistically analyzing many haze-free
outdoor images. Then, an image is dehazed by finding the inverse of
the statistical model's prediction. Fang et al. \cite{fang_variational_2020},
built a variational model and the alternating direction method of
multipliers to do single-image dehazing. Nevertheless, the more complex
the haze, present in real-world imagery, the more challenging it's
for these methods to produce a convincing result.

\subsubsection*{Nighttime}

Even more complex, nighttime haze, has led researchers to devise new
approaches. Zhang et al. \cite{zhang_nighttime_2014,zhang_nighttime_2020},
created a prior, that estimates ambient illumination called maximum-reflectance
prior (MRP), accounting for degradations. Liu et al. \cite{yu_two-branch_2021},
removed noise and light glow using a linear model to decompose an
image into four components. Pei and Lee \cite{pei_nighttime_2012},
used a reference image to map, blue to gray, air-light color, then
applied DCP to dehaze the images. Wang et al. \cite{wang_variational_2022},
exploited the gray-haze line to project, between color spaces, the
concentration of haze in RGB to the Y channel in YUV. And used a variational
structure, estimating the inverted radiance and transmission. Finally,
Liu et al. \cite{liu_single_2022}, developed a variational-decomposition
model to decompose a hazy image into structure, detail, and noise
layers. These systems may fix multiple degradation problems, but introduce
undesirable noise.

\subsection*{Methods Based on Enhancement:}

\subsubsection*{Daytime}

As the field has developed, daytime-dehazing methods have been incorporating
fusion, retinex, and contrast schemes to enhance hazy images. By observing
weather effects in hazy images, Mi et al. \cite{mi_single_2016},
deduced that degradation is concentrated in contrast and color, which
the authors compensated by means of contrast enhancement via a multiscale-gradient
domain. Strategically, Wang et al. \cite{wang_single_2018}, mixed
enhancement and a physical-model, namely multiscale-retinex with color
restoration and transmission-map estimation, improving image quality
lost due to weather degradation. To overcome the limitations of dehazing
with physical models, such as imprecise depth information, Zhu et
al. \cite{zhu_novel_2021}, proposed image fusion, to fuse gamma-corrected
images via pixel-weight maps. Lastly, Liu et al. \cite{liu_joint_2022},
proposed an efficient framework to contrast enhance and fuse exposures.
These methods have yielded good daytime dehazing, but can't be directly
applied to nighttime images.

\subsubsection*{Nighttime}

In similar fashion, other researchers have tackled the night. Utilizing
multiscale fusion, Ancuti et al. \cite{ancuti_night-time_2016,ancuti_effective_2018,ancuti_day_2020},
computed the air light component on different size patches and blended
the multiple images using Laplacian-pyramid decomposition. Finally,
Yu et al. \cite{yu_nighttime_2019}, estimated the transmission map
by combining dark and bright-channel priors. Enhancement-based approaches
are multifaceted, cleverly combining enhancement techniques, achieving
good dehazing. Despite this, they require considerable amount of finetuning,
reducing their generalization.

\subsection*{Methods Based on Deep-Neural Networks}

\subsubsection*{Daytime}

Some approaches with Deep-Neural Networks (DNN) have tried predicting
the transmission map. For example, in DehazeNet \cite{cai_dehazenet_2016},
and MSCNN \cite{ren_single_2016}, neural-networks are used to estimate
the transmission parameters, thereby attaining a haze-free image.
Analogously, Li et al. \cite{li_aod-net_2017}, designed AOD-Net to
learn multiple parameters of the ASM. Similarly, Yang et al. \cite{yang_single_2019},
proposed a region-detection network to predict a transmission map.
Moreover, Zhang et al. \cite{zhang_joint_2020}, estimated a transmission
map, with regularization to control noises. But, there is noise. In
daytime there is need for noise reduction, but more so in nighttime. 

Other approaches have jointly used enhancement and DNNs. Using a network
to dehaze and adaptively enhance, AED-Net, Hovhannisyan et al. \cite{hovhannisyan_aed-net_2022},
applied gray-level dehazing, modified gamma-correction with region
awareness and channel attention. And, combined low and high order
features, removing significant haze. 

The list of methods is extensive, some of the notable ones: GFN \cite{ren_gated_2018},
PDN \cite{yang_proximal_2018}, DRCDN \cite{zhang_drcdn_2020}, NLDN
\cite{zhang_nldn_2020}, FFA-Net \cite{qin_ffa-net_2020}, MSBDN \cite{dong_multi-scale_2020},
DA-Net \cite{shao_domain_2020}, PDR-Net \cite{li_pdr-net_2020},
PSD-Net \cite{chen_psd_2021}, MSAFF-Net \cite{lin_msaff-net_2022},
FSDGN \cite{yu_frequency_2022}, Dehamer \cite{guo_image_2022}, and
AIDED-Net \cite{zhang_aidednet_2023}. 

\begin{figure}[h]
\centering
\subfloat[hazy]{\includegraphics[width=96pt,totalheight=64pt]{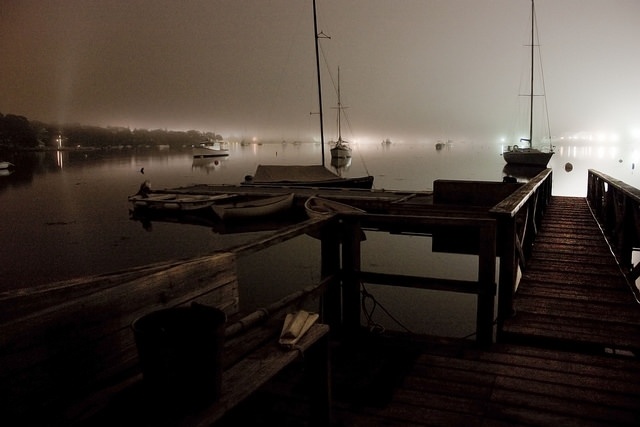}

}\subfloat[FSDGN\cite{yu_frequency_2022}]{\includegraphics[width=96pt,totalheight=64pt]{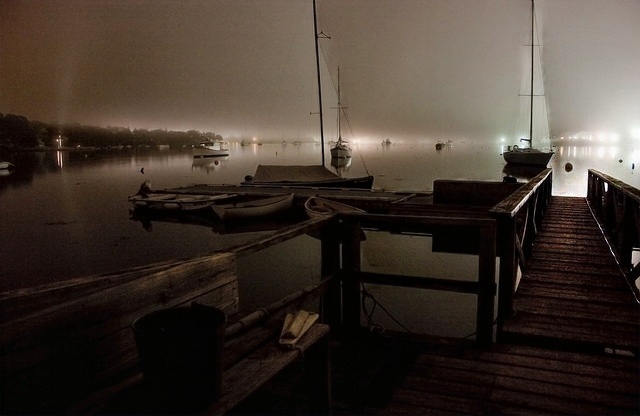}

}

\smallskip{}

\subfloat[Dehamer \cite{guo_image_2022}]{\includegraphics[width=96pt,totalheight=64pt]{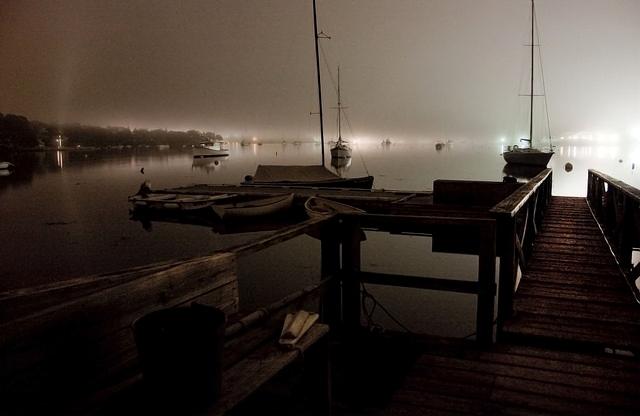}

}\subfloat[NDELS]{\includegraphics[width=96pt,totalheight=64pt]{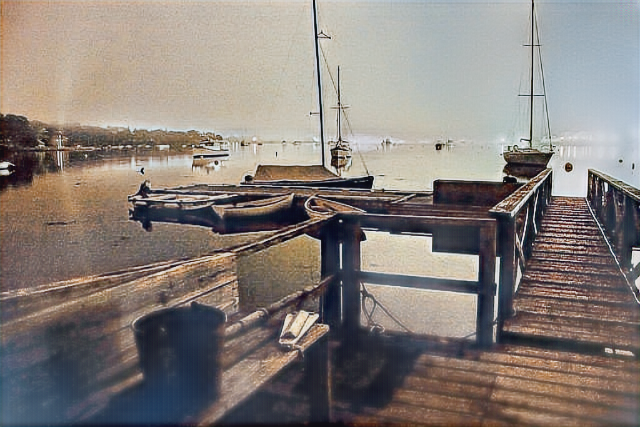}

}

\caption{Performance of daytime dehazing methods on a nighttime hazy image.\textcolor{blue}{\protect\label{fig:Performance-of-daytime-dehazing-on-nighttime-hazy}}}
\end{figure}

Observe the images in Figure \ref{fig:Performance-of-daytime-dehazing-on-nighttime-hazy},
and note the results of the daytime-dehazing methods FSDGN \cite{yu_frequency_2022}
and Dehamer \cite{guo_image_2022}. Compared to NDELS, these two methods
show very little improvement in the visibility of the hazy image.
In general, this class of methods have performed well in the daytime,
but incompletely dehaze in the nighttime, due to non-uniform illumination,
haze, low light, light glow, and other degradation factors.

\subsubsection*{Nighttime}

Another approach has been to isolate light glow. For instance, Koo
and Kim \cite{koo_nighttime_2020}, proposed a glow-decomposition
network to extract glow effects from a hazy image, obtaining atmospheric
light and transmission maps. And, Kuanar et al. \cite{kuanar_night_nodate},
proposed DeGlow, a model to remove glow effects, and DeHaze, a network
for removing haze effects. 

Others have tried separating frequency elements. Yan et al. \cite{yan_nighttime_2020},
build a framework to decompose a grayscale component of a hazy image
into low and high-frequency layers. Then, enhancing the textures in
the high-frequency layer, and removing glow and fog in the lower-frequency
layer. In summary, these nighttime dehazing methods don't provide
consistent quality for real-world image dehazing.

As of late, instance normalization is helping bridge the gap between
daytime and nighttime-dehazing methods. Instance normalization is
showing good results in image restoration tasks. For example, HINet
proposed in \cite{chen_hinet_2021}, a half-instance, normalization
block, replacing batch normalization due to its inefficiency in gathering
accurate feature statistics, as a consequence of using small batches
of data during training. Instead, HINet improves a networks ability
to capture more low-level features by concatenating inputs with their
normalization, which the authors call half-instance normalization.

Narrowing the gap between overexposed and underexposed images by instance
normalizing and mapping into a common space, is the exposure-invariant
space (ENC) module \cite{huang_exposure_2022}. It was designed to
overcome data imbalance, by training with different exposures. But,
there's a worst performing exposure, improved by parameter regularization.
It's plausible, that ENC could benefit daytime and nighttime dehazing.
We could exploit an exposure-invariant space to accomplish dehazing,
then map into the day or night domain. In future work, we will explore
using instance normalization.

This paper proposes a nighttime dehazing, enhancement, and light-suppression
network (NDELS). This network combines (i) a multiscale, attention-guided,
feature-fusion network, designed to brighten poorly illuminated zones,
thereby exposing concealed haze and atmospheric effects, (ii) a modified
Res2Net encoder with residual-channel-attention modules to amplify
low-light regions, eliminate haze and sharpen edges, while mitigating
the impact of excessive illumination, and (iii) a module extending
multiscale Retinex, augmenting color clarity and constancy.

\subsection*{Contributions}

Our paper contributes the following: (a) An end-to-end single-image,
nighttime-dehazing, convolutional, neural network. Its functionality
is independent of external image references, transmission-map estimation,
atmospheric-light analysis, and gathering sequences of images from
a single scene with varying weather effects. (b) A novel approach
of generating training data, crucial for suppressing night-light effects.
(c) The results of detailed computer simulations. And, upon their
analysis, the proposed method outperforms all the state-of-the-art,
single-image, nighttime-dehazing algorithms, namely NDIM \cite{zhang_nighttime_2014},
GS \cite{li_nighttime_2015}, MRP and MRPF \cite{zhang_fast_2017},
OSFD \cite{zhang_nighttime_2020}, FDGCN \cite{koo_nighttime_2020},
GHLP \cite{wang_variational_2022}, and UVD \cite{liu_single_2022}
for real-world images in the NHRW and synthetic NHM datasets \cite{zhang_nighttime_2020}.
Qualitatively, the proposed method does better than non-dehazing,
nighttime-enhancement methods, such as DRSL \cite{sharma_nighttime_2021},
HDRCNN \cite{eilertsen_hdr_2017}, Zero-DCE \cite{guo_zero-reference_2020},
EnlightenGAN \cite{jiang_enlightengan_2019}, and SingleHDR \cite{liu_single-image_2020}.

Table \ref{tab:Comparison-of-state-of-the-art}, summarizes the strengths
and weaknesses of nighttime-dehazing and enhancement methods.

\begin{table*}
\centering
\caption{Comparison of state-of-the-art single-image night dehazing methods.\protect\label{tab:Comparison-of-state-of-the-art}}

\smallskip{}

\centering{}%
\begin{tabular}{c||ccccccccc}
\hline 
 & NDIM\cite{zhang_nighttime_2014} & GS\cite{li_nighttime_2015} & MRP\cite{zhang_fast_2017} & MRPF\cite{zhang_fast_2017} & OSFD\cite{zhang_nighttime_2020} & FDGCN\cite{koo_nighttime_2020} & GHLP\cite{wang_variational_2022} & UVD\cite{liu_single_2022} & NDELS\tabularnewline
\hline 
No assumptions &  &  &  &  & $\checkmark$ & $\checkmark$ &  &  & $\checkmark$\tabularnewline
\hline 
Noise reduction &  &  &  &  &  &  &  & $\checkmark$ & $\checkmark$\tabularnewline
\hline 
Light suppression &  &  &  &  &  &  &  & $\checkmark$ & $\checkmark$\tabularnewline
\hline 
Glow removal &  & $\checkmark$ &  &  &  & $\checkmark$ & $\checkmark$ & $\checkmark$ & $\checkmark$\tabularnewline
\hline 
Non-uniform illumination & $\checkmark$ & $\checkmark$ &  &  & $\checkmark$ & $\checkmark$ & $\checkmark$ & $\checkmark$ & $\checkmark$\tabularnewline
\hline 
Low-light enhancement & $\checkmark$ &  &  &  &  &  &  & $\checkmark$ & $\checkmark$\tabularnewline
\hline 
Nighttime haze removal & $\checkmark$ &  & $\checkmark$ & $\checkmark$ & $\checkmark$ & $\checkmark$ & $\checkmark$ &  & $\checkmark$\tabularnewline
\hline 
Edge enhancement &  &  &  &  &  & $\checkmark$ &  & $\checkmark$ & $\checkmark$\tabularnewline
\hline 
\end{tabular}
\end{table*}

\section*{Proposed Method}

This section provides details on the NDELS architecture, comprised
of three key modules: the low-light, dehazing, and extended-multiscale-retinex.
Furthermore, we discuss the loss functions needed to train the low-light
and dehazing modules.

\subsection*{Overview of the Network Architecture}

Our NDELS network architecture has three essential components:

(i) A multiscale, attention-guided, feature-fusion network, designed
to brighten poorly illuminated zones, exposing concealed haze and
atmospheric effects.

(ii) A modified Res2Net encoder with residual-channel-attention modules
to amplify low-light regions, eliminate haze, and sharpen edges, while
mitigating the impact of excessive illumination.

(iii) A module extending multiscale Retinex, augmenting color clarity
and constancy.

These components are carefully crafted to enhance image quality under
varying illumination conditions, while maintaining the integrity of
the light source.

\begin{figure}
\centering
\includegraphics[width=5cm]{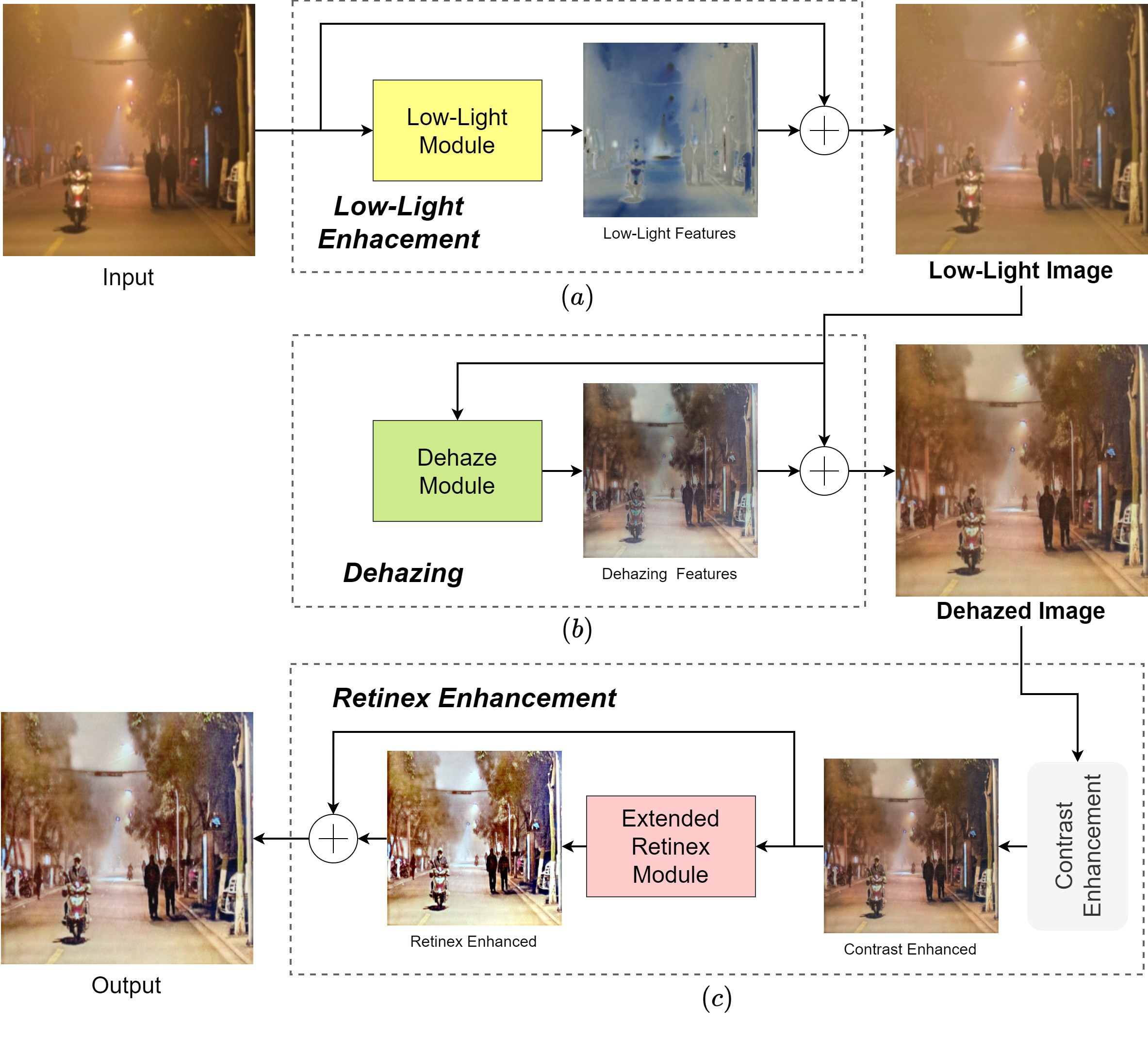}

\caption{High level view of NDELS components. (a) Low-Light Enhancement, (b)
Dehazing, (c) Extended Multiscale Retinex.\protect\label{fig:High-level-view-NDELS}}
\end{figure}

Showcasing the connectivity of the network's modules, Figure \ref{fig:High-level-view-NDELS},
depicts NDELS as a block diagram with high-level details. The low-light
module shines light on the image, but haze is now apparent. The dehazing
module focuses in on the haze features, that are then faded away.
With adjustment of contrast, and an extended-multiscale retinex applied,
original \cite{krishnan_retinex_2022}, the image is considerably
brightened and color enhanced. Our extended-multiscale retinex greatly
improves the visual quality of the image, but overexposes the bright
regions. Thus, we combine the retinex and dehazed images, producing
an image of higher quality.

\begin{figure*}
\centering
\begin{centering}
\includegraphics[scale=0.15]{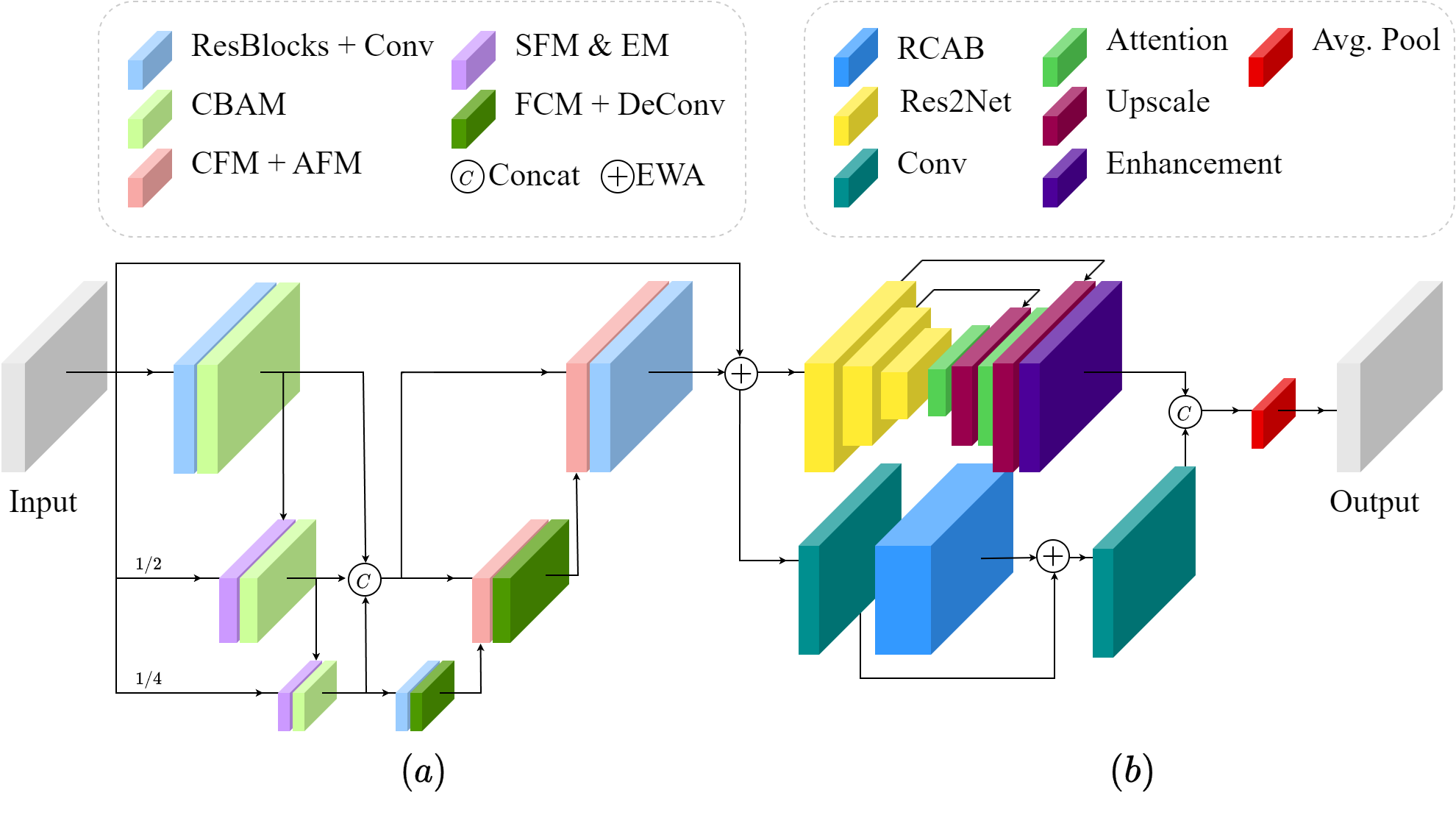}
\par\end{centering}
\caption{Network architecture of NDELS. (a) Low-Light Module and (b) Dehazing
Module.\protect\label{fig:Low-light-and-Dehazing}}
\end{figure*}

\emph{Low-Light Module} (LLM): The low-light module, part (a) of Figure
\ref{fig:Low-light-and-Dehazing}, takes in a low-light image with
haze. We have implemented our own version of the Attention-Guided,
Multiscale, Fusion Network (AGMFN) \cite{cui_attention-guided_2022}.
The key components are the resblocks, convolutional-block-attention
(CBAM), cascade-fusion (CFM), attention-fusion (AFM), feature-enhancement
(FEM), and feature-calibration modules (FCM). The input image passes
through a set of resblocks and convolutional layers, then through
the CBAM. It guides the low-light module in enhancing poorly lit regions,
and restricting the over-enhancement of bright areas. 

The two remaining encoder-decoder branches pass a downscaled image
through shallow feature (SFM) and FEM modules. The SFM convolves the
image through a series of low-order kernels to capture low-level features.
Given the encoded features, the FEM combines them with those of the
SFM. Fusing the low-level and multiscale-feature maps, the AFM uses
global-average pooling to amplify channel information and strengthen
semantic dependencies. The CFM concatenates the encoded data from
the three scales to compensate for the loss of information due to
the downscaling. The remaining feature-encoded information is fed
back into the adjacent higher-scale branch through decoding via deconvolution,
and linked by the FCM. Improving degradation and detail enhancement
is accomplished by the FCM, which passes the important lower-scale
features to the AFM.

\emph{Dehazing Module} (DHM): To tackle the issue of haze present
in hazy nighttime images, we employ the dehazing module proposed in
\cite{yu_two-branch_2021}. Although this network is meant to dehaze
daytime images, we prove that by using a low-light module to enhance
the low-light regions and the haze, it can be trained to dehaze nighttime
images with the appropriate training data. Part (b) of Figure \ref{fig:Low-light-and-Dehazing},
depicts the dehazing module with two branches: the upper pretrained
and the lower trained branches. 

The upper, consists of a Res2Net encoder-decoder module \cite{gao_res2net_2019}.
It uses feature-fusion attention \cite{qin_ffa-net_2020} as the attention
block, PixelShuffle \cite{shi_real-time_2016} for upscaling, and
an enhancement module \cite{qu_enhanced_2019}. ImageNet weights are
loaded during training, allowing the network to extract more robust
features rather than randomly initializing the weights. The lower
branch captures finer details, avoiding rescaling, and adjusts the
network to the dehazed image. The branch comprises residual-channel-attention
blocks (RCAB) \cite{zhang_image_2018}. Each residual-channel-attention
module (RCAM) is composed of RCABs with a long skip connection, reducing
the gradient vanishing problem \cite{pascanu_difficulty_2012}. The
two branches are concatenated, passing through a convolutional layer
and $tanh$ activation.

\emph{Discriminator}: We use a discriminator network-based generative
model to differentiate between real and generated data, which can
improve the dehazing module's overall performance and its ability
to produce more robust features that resemble real data, making the
generated data lifelike \cite{yu_two-branch_2021,ledig_photo-realistic_2016}.
The following structure is adopted: The network starts by taking an
image as input, which is then scaled up to 512 output channels using
multiple convolutional layers with a kernel size of $3\times3$. These
convolutional layers are followed by batch normalization and leaky-ReLU
activation, stabilizing the training process and avoiding vanishing
gradients---the negative activation slope allows small gradients
to flow through during training. Then adaptive-average pooling is
applied to the feature maps, reducing spatial dimensions and improving
translation robustness. Two convolutional layers follow, with kernel
size $1\times1$, and output channels of 1024 and 1. These layers
reduce the number of channels to one. Finally, a sigmoid activation
normalizes the probability, that an image is real or synthetic.

\emph{Extended-Multiscale-Retinex} (EMSR): The multiscale-retinex
method \cite{petro_multiscale_2014}, is a human-perception based,
color-image, enhancement algorithm. EMSR utilizes the Fast-Fourier-Transform
(FFT) to filter each color channel, using Gaussian kernels with scales
of 5, 130, and 255. Low and high pixel intensities are analyzed, keeping
the most frequent, with frequency less than 10\% the frequency of
pixels with 0 intensity. The color channels are combined yielding
a contrast and brightness-improved image, visually more appealing,
easier to interpret, and better color constancy.

\emph{Loss Functions}: To train the LLM, we compute the content loss
calculated using the $L_{1}$ loss and multiscale-frequency-domain
loss (MSFD) \cite{cui_attention-guided_2022,li_luminance-aware_2021}.
The total loss for the LLM is the linear combination of the content
loss and MSFD loss given by
\[
L_{\text{total}}=\gamma_{1}L_{\text{content}}+\gamma_{2}L_{\text{MSFD}},
\]
 where the coefficients $\gamma_{1}$and $\gamma_{2}$ are hyperparameters.
To train the DHM, we compute the perceptual\cite{johnson_perceptual_2016,simonyan_very_2014},
smoothness $L_{1}$ \cite{girshick_fast_2015}, multiscale, structural-similarity-index
measure (MS-SSIM) \cite{trongtirakul_single_2020}, and adversarial
losses \cite{ledig_photo-realistic_2016}. The total loss for the
DHM is the linear combination of the four losses defined as:
\[
L_{\text{total}}=\gamma_{1}L_{l1}+\gamma_{2}L_{\text{MS-SSIM}}+\gamma_{3}L_{\text{perc}}+\gamma_{4}L_{\text{adv}},
\]
where the coefficients $\gamma_{1},\gamma_{2},\gamma_{3}$and $\gamma_{4}$
are hyperparameters.

We believe that the MSFD losses are better suited for the type of
task that requires for the restoration of hidden or degradated details
in an image, which is typical of nighttime or low-light imagery. Additionally,
this loss helps reduce and suppress the light effects. For the DHM,
the smoothness $L_{1}$, perceptual, and MS-SSIM, ensure that the
dehazed image is close to the target perceptually and structurally,
based on human-visual perception and hidden features from a neural
network. As for the adversarial loss, it ensures the final dehazed
image is realistic.

\section*{Experiments}

This section provides quantitative and qualitative comparisons of
NDELS with state-of-the-art, nighttime-dehazing algorithms.

\subsection*{Datasets}

We choose the following four datasets to train, validate, and test
our proposed method.

\textbf{Nighttime Hazy Middlebury (NHM)}: The NHM dataset \cite{zhang_nighttime_2020},
contains 350 synthetic images, taken at night, with low, medium, and
high-haze densities. With the dataset we quantitatively compare our
method with state-of-the-art methods available at the time of this
writing. In addition, we select five samples of varying haze density,
and measure the quality relative to the ground truth.

\textbf{Transient-Attributes Dataset (TA)}: The TA dataset \cite{laffont_transient_2014},
contains 8,571 images from 101 webcams with 40 attribute labels. We
select images with dark and bright attributes, such that the attribute
confidence is at least 80\%.

\textbf{Nighttime Hazy Real-World (NHRW)}: The NHRW dataset \cite{zhang_nighttime_2020},
contains 150 real-world, nighttime images with low, medium, and high-haze
densities. Qualitatively, with this dataset we compare our method
with leading others, measuring their effectiveness in dehazing and
enhancing nighttime images.

\textbf{Light-Effects Dataset (LE)}: The LE dataset \cite{sharma_nighttime_2021},
contains 501 nighttime, haze-free images having light glare, glow,
and flood lights. We compare, qualitatively with the dataset, our
model’s light-suppressing capabilities to other state-of-the-art methods
for nighttime enhancement.

\begin{table}
\centering
\caption{Breakdown of training and validation data generated from a subset
of the TA dataset.\protect\label{tab:Breakdown-of-training}}

\centering{}%
\begin{tabular}{c||c|c|c|c}
\hline 
Task & Cameras & Bright & Dark & Generated Pairs\tabularnewline
\hline 
Training & 52 & 550 & 331 & 3056\tabularnewline
\hline 
Validation & 9 & 60 & 76 & 385\tabularnewline
\hline 
\end{tabular}
\end{table}

\subsection*{Implementation Details}

\begin{figure}
\centering
\begin{centering}
\subfloat[bright]{\includegraphics[scale=0.1]{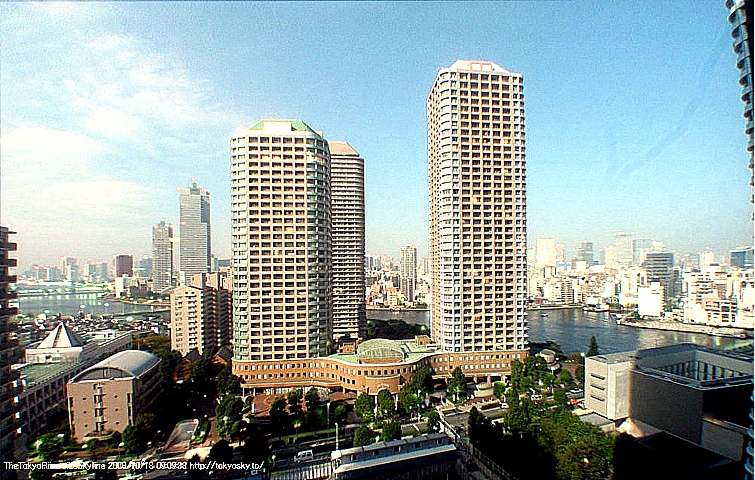}

}\subfloat[bright-haze]{\includegraphics[scale=0.1]{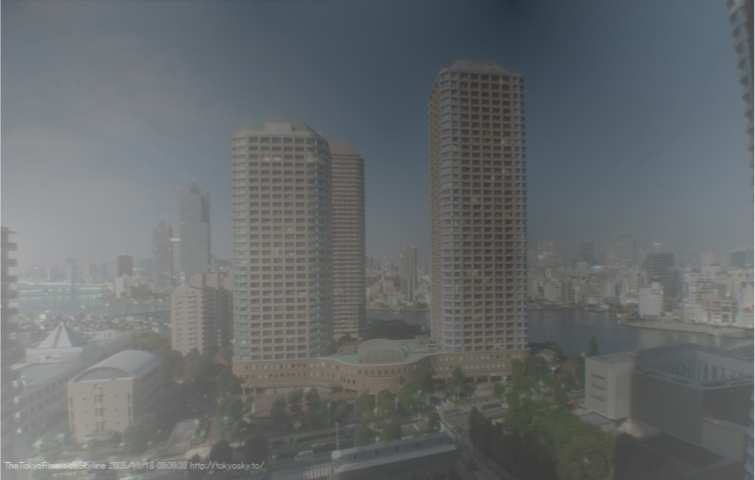}

}\subfloat[dark-haze]{\includegraphics[scale=0.1]{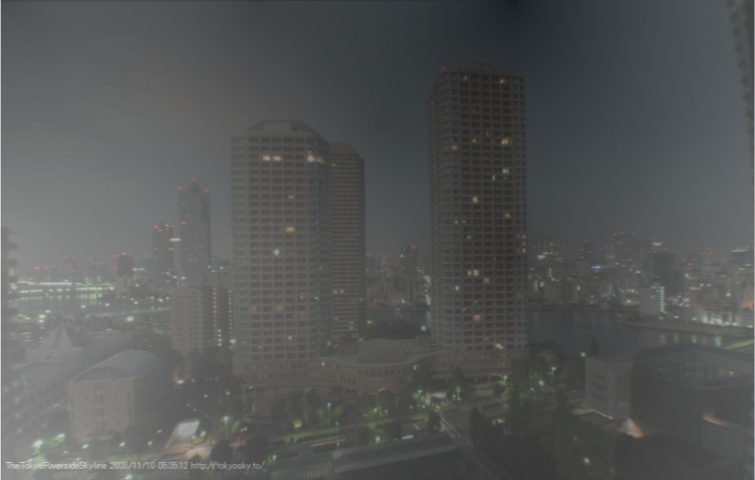}

}
\par\end{centering}
\caption{Synthetic training triplet generated from the TA dataset.\protect\label{fig:Example-training-triplet}}
\end{figure}

To enable the network to effectively learn the processes of dehazing
and low-light enhancement, we have devised a strategy that utilizes
both night and day imagery. The implementation involves the following
steps:
\begin{itemize}
\item Data Gathering:
\begin{itemize}
\item Utilizing camera data from the TA dataset.
\item Alternatively, recommending the use of fixed-view, internet webcams
capturing day and night outdoor scenes with elements such as vehicles,
buildings, and night lights. For efficient data gathering, we suggest
capturing one image every 12 hours.
\end{itemize}
\item Image Composition:
\begin{itemize}
\item Obtaining an image pair, bright image $(B)$ and dark image $(D)$,
from the TA dataset.
\item Generating a new bright image $(B')$ through linear combination with
the dark image $(D)$: $B'=0.7B+0.3D$.
\item Generating a new dark image $(D')$: $D'=0.3B+0.7D$.
\end{itemize}
\item Synthetic-Haze Addition:
\begin{itemize}
\item Adding random synthetic haze $(H)$ to each pair $B'$ and $D'$ using
functions from \cite{buslaev_albumentations_2020}.
\item Resulting in images: bright, bright-hazy, and dark-hazy ($B',B'+H,D'+H$,
respectively).
\end{itemize}
\item Image Enhancement:
\begin{itemize}
\item Sharpening and contrast enhancing the bright image $(B')$ further
by clipping the bottom and top 5\% of intensity values in each color
channel.
\end{itemize}
\end{itemize}
Observe the presence of lights inside windows in the bright image
$(B')$ in Figure \ref{fig:Example-training-triplet}, which constrains
the low-light enhancement and dehazing modules from over enhancing
light sources. During training, each training image is resized to
$512\times256$ pixels and randomly cropped to snippets of $256\times256$
pixels, randomly rotated and flipped. The validation images are resized
to $512\times256$ pixels. The low-light enhancement module learns
to map images that are dark-hazy to bright-hazy. And, the dehazing
module learns to map images that are bright-hazy to bright-dehazed.

The low-light, enhancement module uses the total-loss function described
in the previous section. Reiterating, the total loss is the sum of
the content loss and the MSFD loss. In addition, the dehazing module
uses a perceptual loss through VGG16 \cite{simonyan_very_2014}. Along
with MS-SSIM, smooth, and adversarial losses. During training, the
learning rate is initialized to $10^{-4}$, and decreased every 14
epochs by 10, for 42 epochs. Approximately 9 hours of training are
needed when performed on an NVIDIA RTX A6000 with PyTorch 1.13.

\subsection*{Qualitative Evaluation}

\begin{figure*}
\centering
\begin{centering}
\subfloat[]{\includegraphics[scale=0.75]{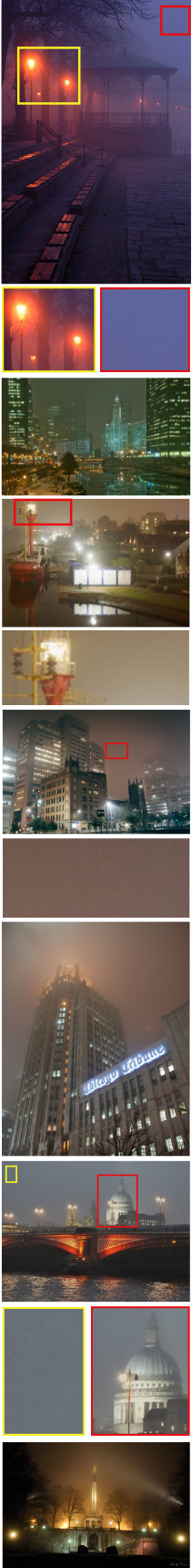}{\scriptsize}}\subfloat[]{\includegraphics[scale=0.75]{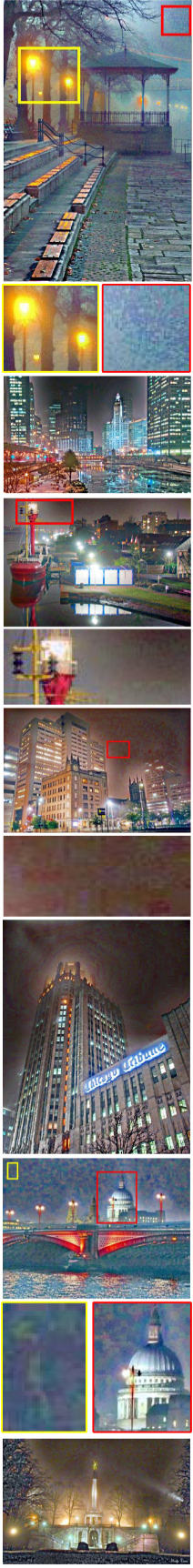}

}\subfloat[]{\includegraphics[scale=0.75]{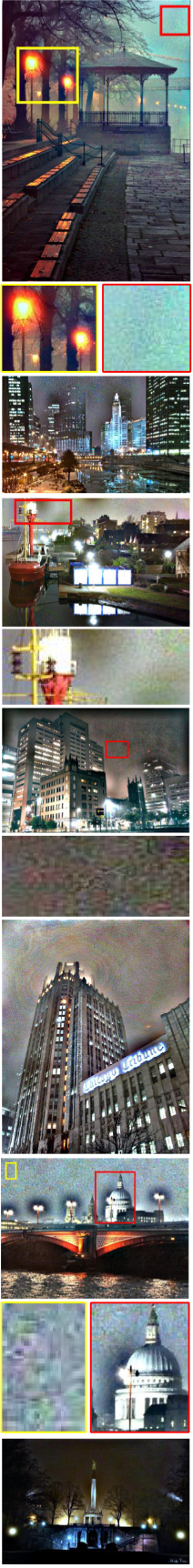}

}\subfloat[]{\includegraphics[scale=0.75]{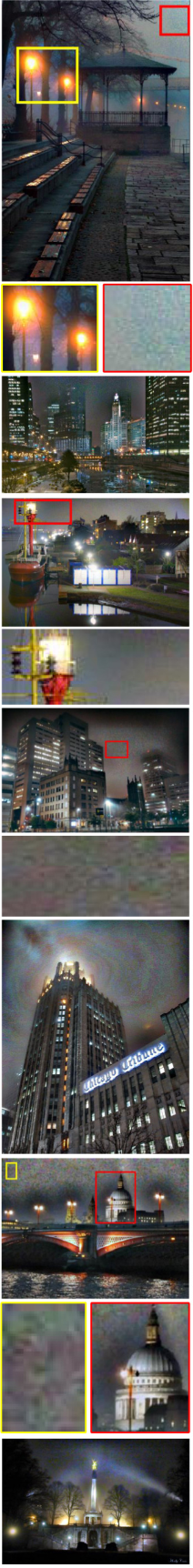}

}\subfloat[]{\includegraphics[scale=0.75]{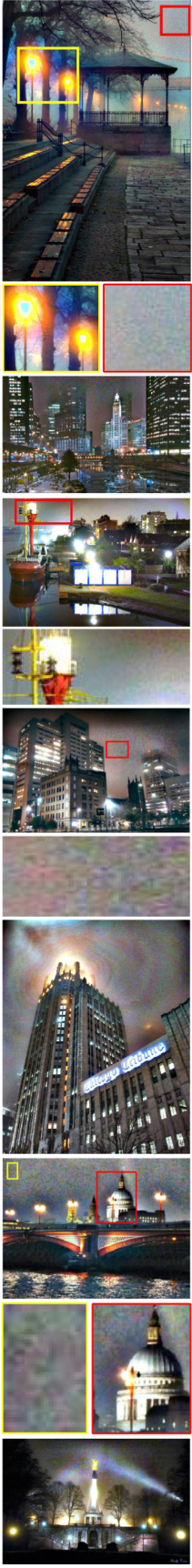}

}\subfloat[]{\includegraphics[scale=0.75]{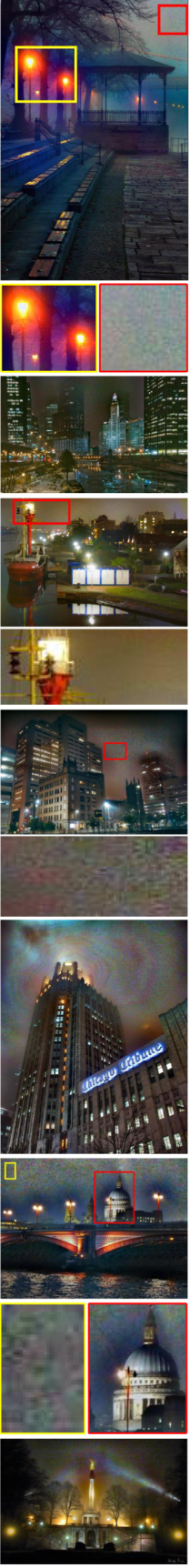}

}\subfloat[]{\includegraphics[scale=0.75]{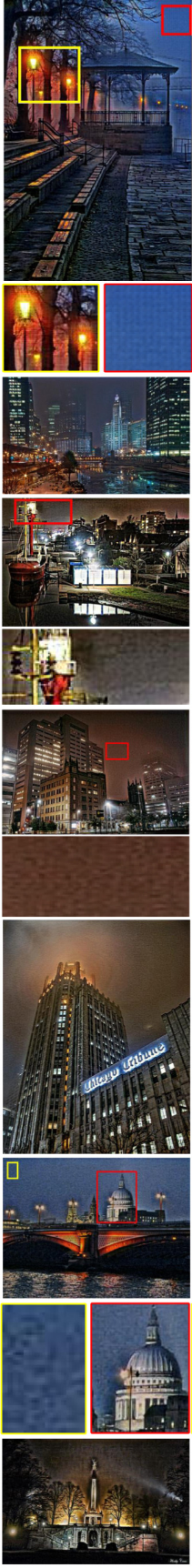}

}\subfloat[]{\includegraphics[scale=0.75]{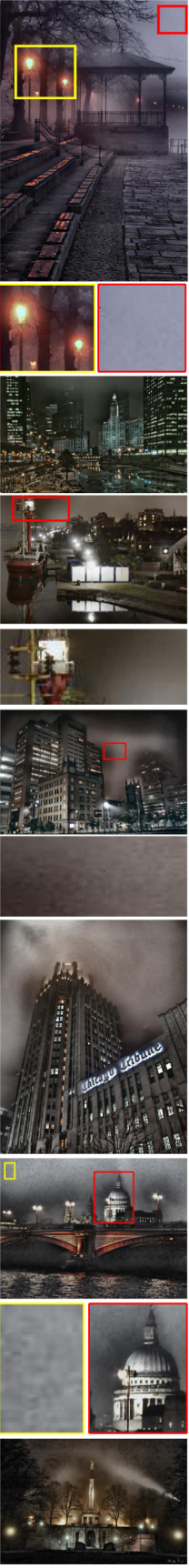}

}\subfloat[]{\includegraphics[scale=0.75]{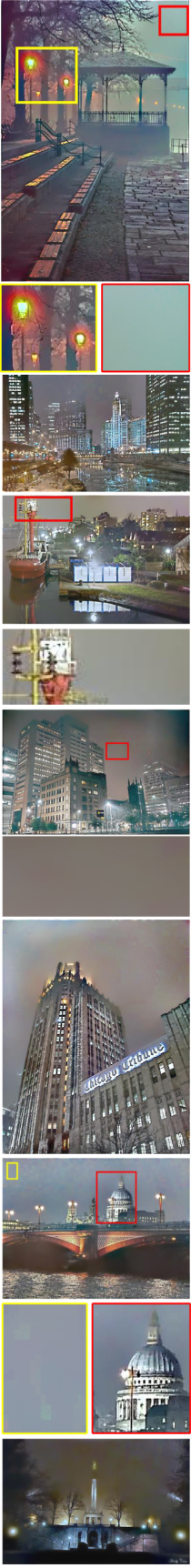}

}\subfloat[]{\includegraphics[scale=0.75]{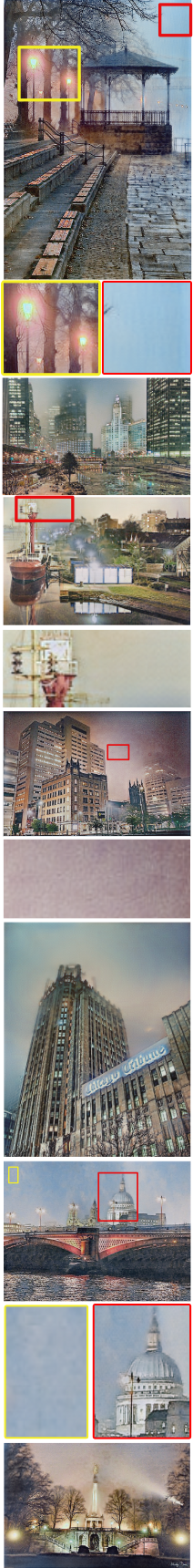}

}
\par\end{centering}
\caption{Qualitative comparison of single-image, nighttime, dehazing algorithms
on real-world images from the NHRW dataset. (a) inputs, (b) NDIM \cite{zhang_nighttime_2014},
(c) GS \cite{li_nighttime_2015}, (d) MRP and (e) MRPF \cite{zhang_fast_2017},
(f) OSFD \cite{zhang_nighttime_2020}, (g) FDGCN \cite{koo_nighttime_2020},
(h) GHLP \cite{wang_variational_2022}, (i) UVD \cite{liu_single_2022},
and (j) NDELS. \protect\label{fig:Qualitative-comparison-of}}
\end{figure*}

In Figure \ref{fig:Qualitative-comparison-of}, we show a qualitative
comparison of NDELS with other algorithms. In the first image, all
methods reduce haze. The brightest appears to be (b) NDIM \cite{zhang_nighttime_2014},
but at the cost of high noise and further brightening the bright lamps.
Methods (g) FDGCN \cite{yan_nighttime_2020} and (h) GHLP \cite{wang_variational_2022}
barely illuminate the input, but don't worsen the light effects. However,
visible haze remains. The second best, qualitatively is (i) UVD \cite{liu_single_2022},
which produces a good image, but with blur and some haze remaining.
In comparison, (j) NDELS has a sharp, bright, and dehazed image, with
light-effects suppression.

\begin{figure*}
\centering
\begin{centering}
\subfloat[]{\includegraphics[scale=0.75]{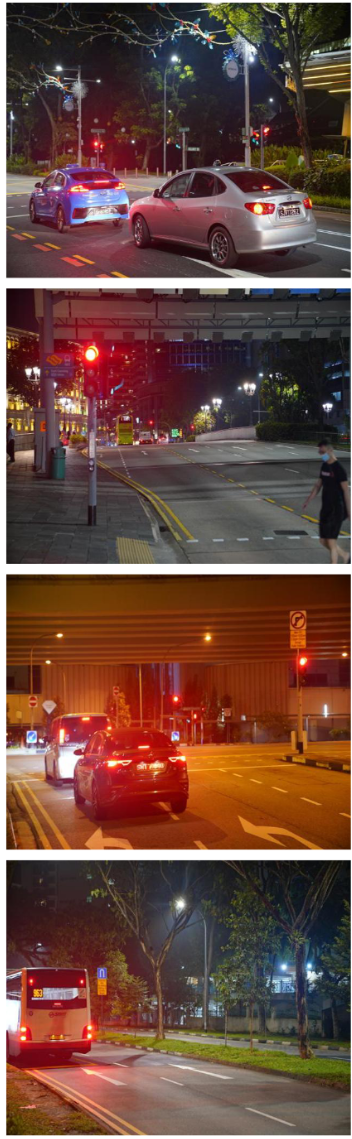}

}\subfloat[]{\includegraphics[scale=0.75]{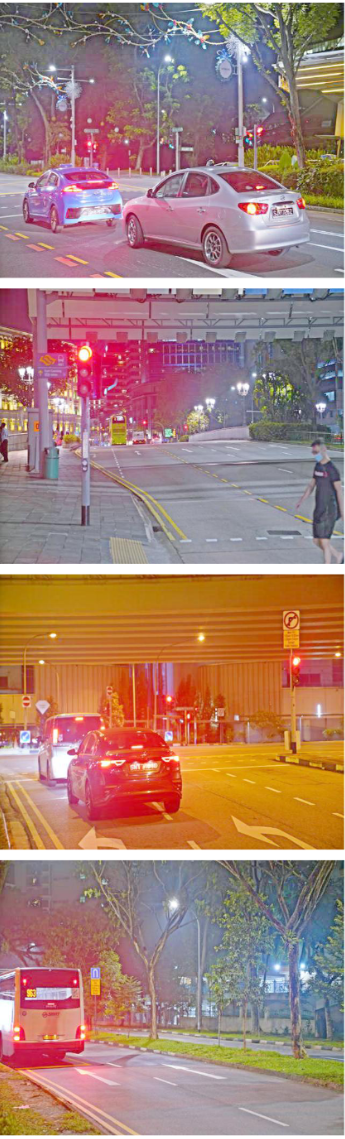}

}\subfloat[]{\includegraphics[scale=0.75]{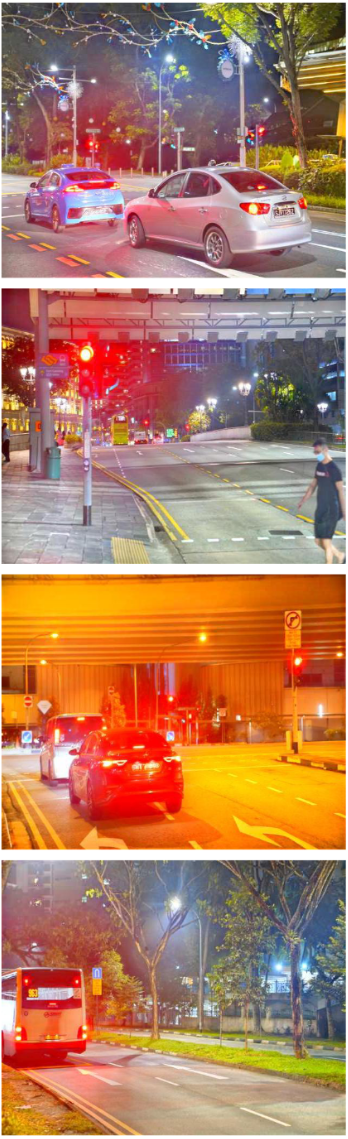}

}\subfloat[]{\includegraphics[scale=0.75]{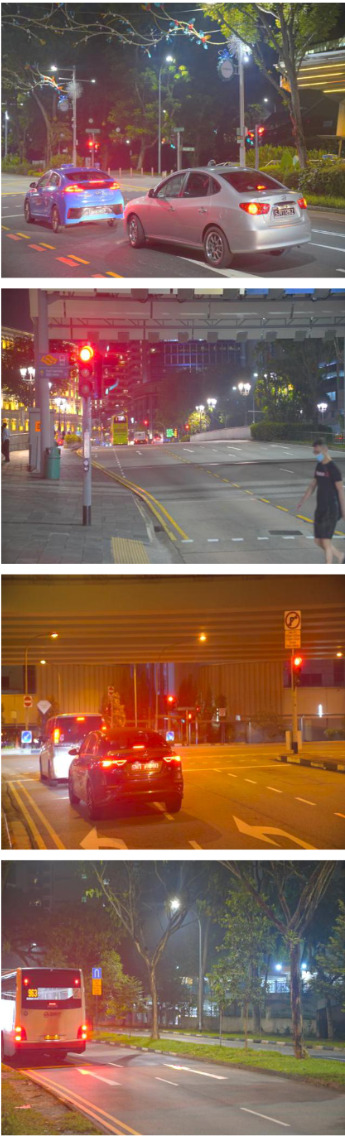}

}\subfloat[]{\includegraphics[scale=0.75]{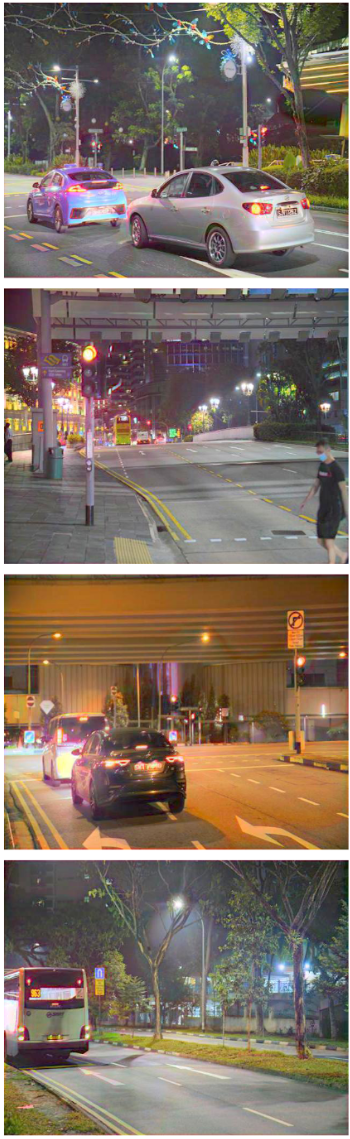}

}\subfloat[]{\includegraphics[scale=0.75]{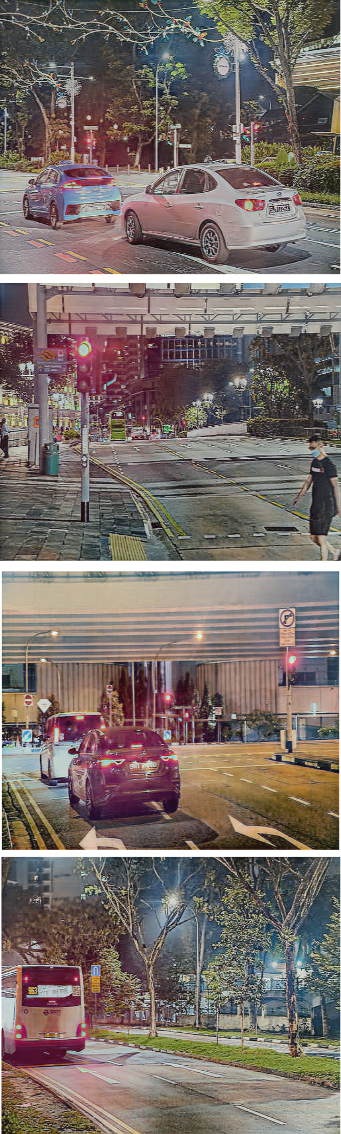}

}
\par\end{centering}
\caption{Qualitatively comparing state-of-the-art nighttime-image, enhancement
methods for images with glow, glare, and floodlight light effects.
(a) inputs, (b) Zero-DCE \cite{guo_zero-reference_2020}, (c) EnlightenGAN
\cite{jiang_enlightengan_2019}, (d) SingleHDR \cite{liu_single-image_2020},
(e) DRSL \cite{sharma_nighttime_2021}, and (f) NDELS. \protect\label{fig:Qualitative-night-image-enhancement}}
\end{figure*}

Shown in Figure \ref{fig:Qualitative-night-image-enhancement}, is
a qualitative comparison of night-enhancement algorithms with NDELS.
The DRSL \cite{sharma_nighttime_2021} algorithm does a good job of
removing light effects, but doesn't enhance all low-light areas. Others,
such as Zero-DCE \cite{guo_zero-reference_2020}, EnlightenGAN \cite{jiang_enlightengan_2019},
and SingleHDR \cite{liu_single-image_2020}, distort the input image
by expanding the red-color lights and adding some haze to the image.
Remaining methods blur and over enhance bright regions or increase
haze. But, NDELS successfully suppresses light effects, glow, glare,
and flood lights, sharpening and enhancing low-light areas, with the
added benefit of removing nighttime haze.

\subsection*{Quantitative Evaluation}

\begin{figure*}
\centering
\begin{centering}
\subfloat[]{\includegraphics[scale=0.75]{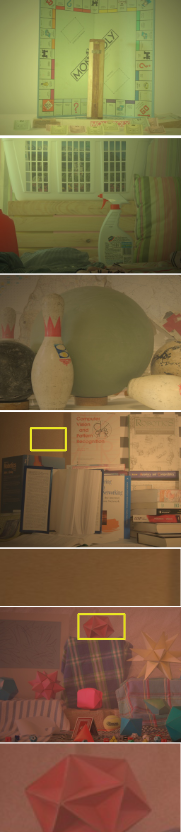}

}\subfloat[]{\includegraphics[scale=0.75]{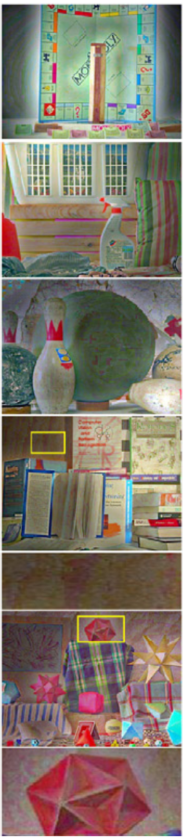}

}\subfloat[]{\includegraphics[scale=0.75]{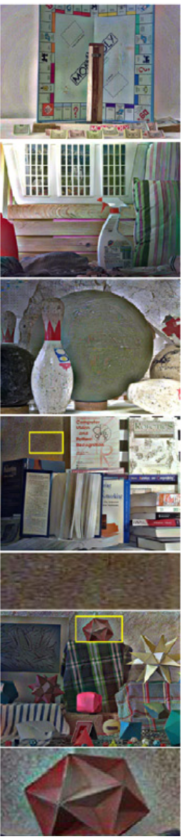}

}\subfloat[]{\includegraphics[scale=0.75]{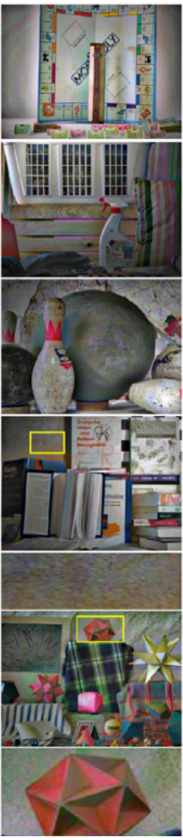}

}\subfloat[]{\includegraphics[scale=0.75]{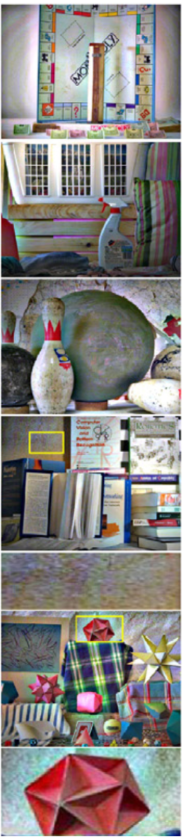}

}\subfloat[]{\includegraphics[scale=0.75]{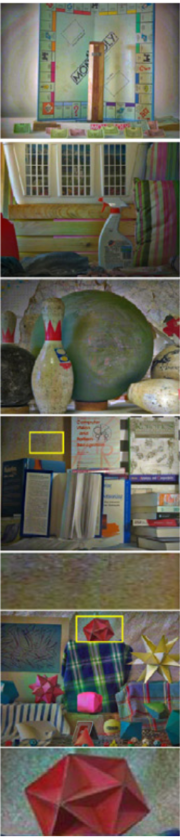}

}\subfloat[]{\includegraphics[scale=0.75]{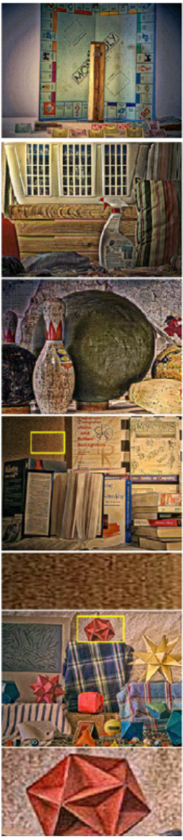}

}\subfloat[]{\includegraphics[scale=0.75]{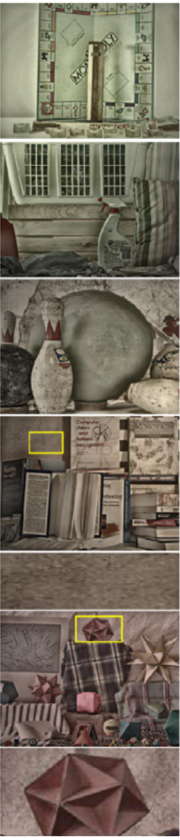}

}\subfloat[]{\includegraphics[scale=0.75]{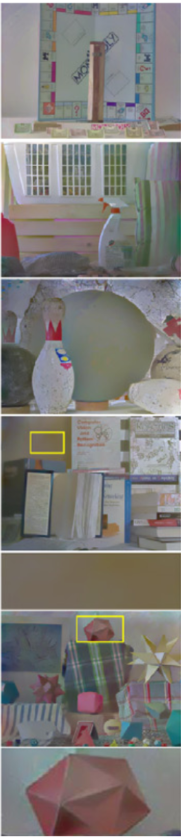}

}\subfloat[]{\includegraphics[scale=0.75]{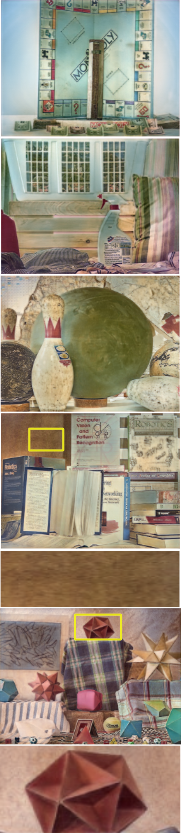}

}\subfloat[]{\includegraphics[scale=0.75]{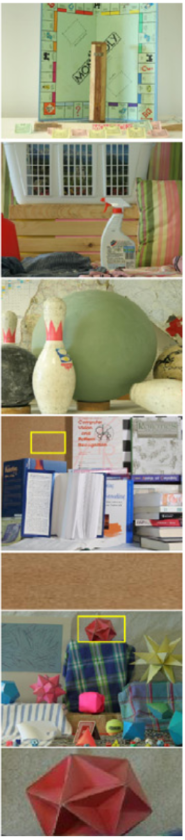}

}
\par\end{centering}
\caption{Quantitative comparison of single-image, nighttime-dehazing algorithms
on selected images from the synthetic NHM dataset. (a) inputs, (b)
NDIM \cite{zhang_nighttime_2014}, (c) GS \cite{li_nighttime_2015},
(d) MRP and (e) MRPF \cite{zhang_fast_2017}, (f) OSFD \cite{zhang_nighttime_2020},
(g) FDGCN \cite{koo_nighttime_2020}, (h) GHLP \cite{wang_variational_2022},
(i) UVD \cite{liu_single_2022}, (j) NDELS, and (k) GT. \protect\label{fig:Quantitative-comparison-of}}
\end{figure*}

In our evaluation to quantify the NDELS method, we adopt the peak-signal-to-noise
ratio (PSNR) and the structural-similarity-index measure (SSIM) \cite{wang_image_2004}.
Five representative images from the NHM dataset are shown in Figure
\ref{fig:Quantitative-comparison-of}. The scores of these images
are given in the top rows of Table \ref{tab:Quantitative_summary}.
The table demonstrates that NDELS consistently scores higher for both
metrics. The last two rows are for the entire NHM dataset. NDELS has
a higher PSNR and SSIM than all the other methods. These quantitative
results are for NDELS without the EMSR module. We find that the module
works best on real-world images.

\begin{table*}
\centering
\caption{PSNR (top) and SSIM (bottom) scores for a subset and full NHM dataset.\protect\label{tab:Quantitative_summary}}

\centering{}%
\begin{tabular}{c||c|c|c|c|c|c|c|c|c|c}
\hline 
\multirow{13}{*}{Subset} & Image & NDIM\cite{zhang_nighttime_2014} & GS\cite{li_nighttime_2015} & MRP\cite{zhang_fast_2017} & MRPF\cite{zhang_fast_2017} & OSFD\cite{zhang_nighttime_2020} & FDGCN\cite{koo_nighttime_2020} & GHLP\cite{wang_variational_2022} & UVD\cite{liu_single_2022} & NDELS\tabularnewline
\cline{2-11}
 & \multirow{2}{*}{1} & 11.0509 & 14.8753 & 12.2937 & 13.9119 & 12.7518 & 9.5976 & 14.5704 & 15.5548 & \textbf{16.2492}\tabularnewline
 &  & 0.6215 & 0.6767 & 0.6868 & 0.6780 & 0.6913 & 0.4908 & 0.7761 & \textbf{0.8193} & 0.7996\tabularnewline
\cline{2-11}
 & \multirow{2}{*}{2} & 16.1221 & 13.9817 & 14.7126 & 14.617 & 14.6873 & 13.3735 & 16.0579 & 16.6188 & \textbf{16.7865}\tabularnewline
 &  & 0.6927 & 0.6782 & 0.6892 & 0.6818 & 0.7084 & 0.4087 & 0.7494 & 0.7414 & \textbf{0.7529}\tabularnewline
\cline{2-11}
 & \multirow{2}{*}{3} & 12.6267 & 12.9039 & 12.0158 & 13.7679 & 12.0900 & 10.2708 & 13.7826 & 14.7216 & \textbf{17.4966}\tabularnewline
 &  & 0.6080 & 0.5818 & 0.6177 & 0.6171 & 0.6331 & 0.2611 & 0.6937 & 0.7387 & \textbf{0.7653}\tabularnewline
\cline{2-11}
 & \multirow{2}{*}{4} & 13.1032 & 13.3063 & 12.1928 & 14.6411 & 11.8457 & 10.8901 & 11.6252 & 14.8185 & \textbf{17.4809}\tabularnewline
 &  & 0.6422 & 0.6626 & 0.6465 & 0.6654 & 0.6562 & 0.4491 & 0.6709 & 0.7007 & \textbf{0.7492}\tabularnewline
\cline{2-11}
 & \multirow{2}{*}{5} & 14.0529 & 14.8528 & 16.0453 & 15.0434 & 16.0152 & 14.5347 & 16.1686 & \textbf{17.2239} & 16.6297\tabularnewline
 &  & 0.6225 & 0.6640 & 0.6996 & 0.6366 & 0.7078 & 0.4501 & 0.7185 & 0.7233 & \textbf{0.7395}\tabularnewline
\cline{2-11}
 & \multirow{2}{*}{Avg} & 13.3912 & 13.9840 & 13.4520 & 14.3963 & 13.4780 & 11.7333 & 14.4409 & 15.7875 & \textbf{16.9286}\tabularnewline
 &  & 0.6374 & 0.6527 & 0.6680 & 0.6558 & 0.6794 & 0.4120 & 0.7217 & 0.7447 & \textbf{0.7613}\tabularnewline
\hline 
\hline 
\multirow{2}{*}{Full} & \multirow{2}{*}{-} & 12.4924 & 11.8963 & 12.9928 & 13.1847 & 13.3027 & 11.5466 & 13.1523 & 13.4679 & \textbf{14.6458}\tabularnewline
 &  & 0.5752 & 0.5899 & 0.6299 & 0.6164 & 0.6435 & 0.3684 & 0.6735 & 0.6630 & \textbf{0.7035}\tabularnewline
\hline 
\end{tabular}
\end{table*}

\subsection*{Object Detection}

\begin{figure}[h]
\centering
\subfloat[no dehazing]{\includegraphics[viewport=0bp 0bp 640bp 480bp,width=2.8cm,height=2cm]{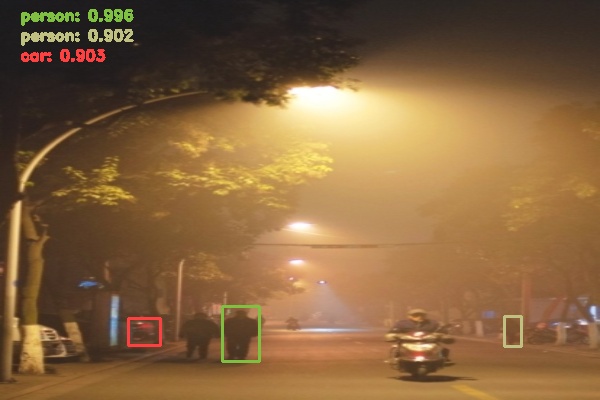}

}\negthickspace{}\negthickspace{}\subfloat[MRP\cite{zhang_fast_2017}]{\includegraphics[viewport=0bp 0bp 640bp 480bp,width=2.8cm,height=2cm]{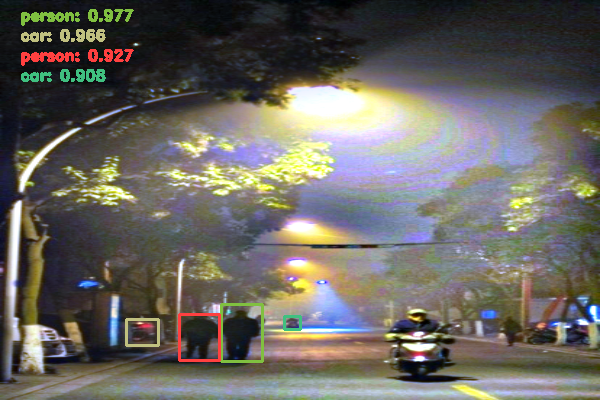}

}\negthickspace{}\negthickspace{}\subfloat[MRPF\cite{zhang_fast_2017}]{\includegraphics[viewport=0bp 0bp 640bp 480bp,width=2.8cm,height=2cm]{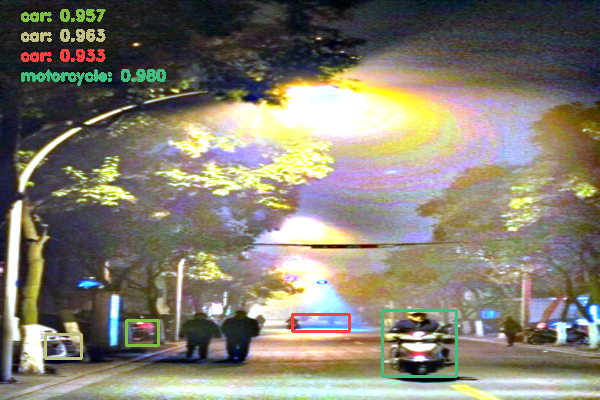}

}\vspace{-0.5cm}

\subfloat[OSFD\cite{zhang_nighttime_2020}]{\includegraphics[viewport=0bp 0bp 640bp 480bp,width=2.8cm,height=2cm]{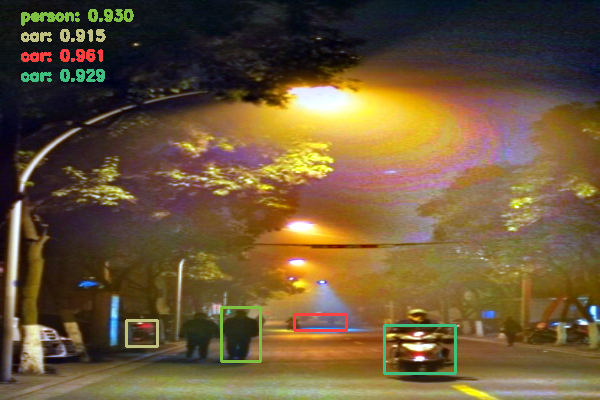}}\negthickspace{}\subfloat[NDELS]{\includegraphics[viewport=0bp 0bp 640bp 480bp,width=2.8cm,height=2cm]{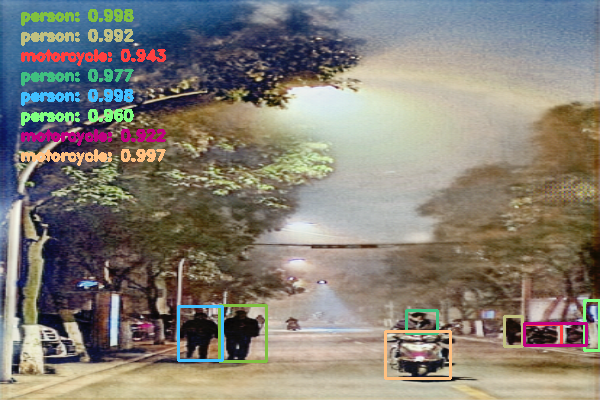}}

\caption{DETR object-detection model applied to nighttime-dehazing algorithms.\textcolor{blue}{{}
\protect\label{fig:DETR-object-detection-model}}}
\end{figure}

In order to assess the real-world applicability of NDELS, particularly
in the context of object detection, we selected a test image from
the NHRW dataset. Subsequently, we process the image through our network
and, for comparative analysis, through an established object detection
network, DETR \cite{carion_end--end_2020}. As illustrated in Figure
\ref{fig:DETR-object-detection-model}, the application of NDELS results
in a significant improvement in both accuracy and the number of objects
detected, outperforming MRP, MRPF, and OSFD. It's noteworthy that
certain methods, such as the GHLP, couldn't be easily compared due
to the unavailability of public code. Even when implementing GHLP
based on information provided in the published paper, the results
were unfavorable.

\subsection*{Subjective Study}

\begin{figure}[h]
\centering
\includegraphics[width=6cm]{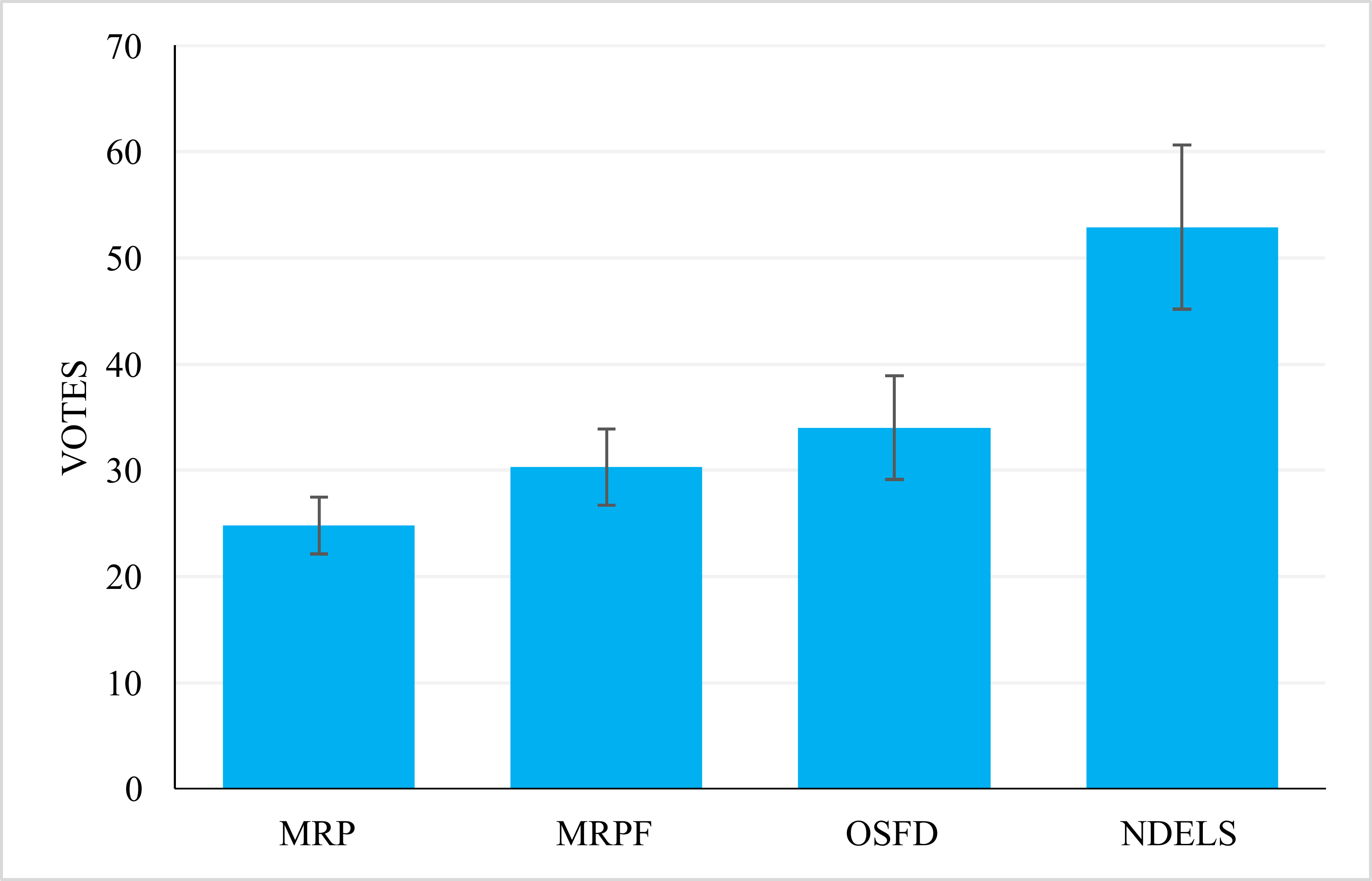}

\caption{Survey on image quality of MRP\cite{zhang_fast_2017}, MRPF\cite{zhang_fast_2017},
OSFD\cite{zhang_nighttime_2020}, and NDELS. Votes are average number
of people that chose a model's output image.\protect\label{fig:Survey-on-image}}
\end{figure}

The results of the questionnaire align with expectations. We randomly
selected ten images from the NHRW dataset and surveyed 142 individuals.
Participants were tasked with selecting the image that demonstrated
the highest clarity, visibility, and detail. Each question featured
one original image and the outputs of four models. As depicted in
Figure \ref{fig:Survey-on-image}, our analysis reveals that, on average,
NDELS outperforms the other models by at least $18.9\pm7.7$ votes.
In conclusion, the survey findings imply that NDELS generally produces
a subjectively better image.

\subsection*{Ablation Study}

\begin{figure*}
\centering
\begin{centering}
\includegraphics[width=1\textwidth]{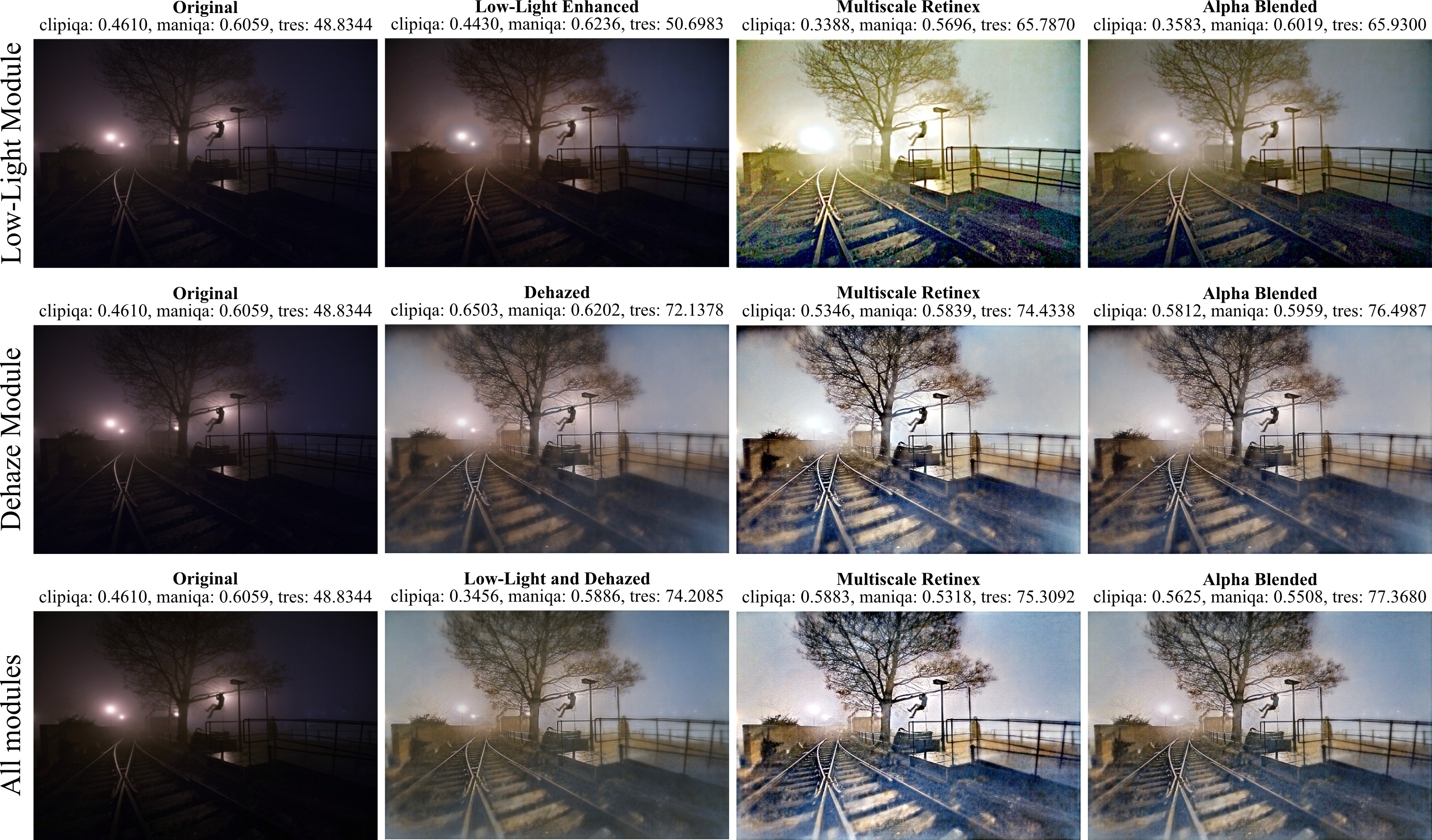}
\par\end{centering}
\caption{Ablation study testing NDELS modules on a real-world nighttime hazy
image from the NHRW dataset.\protect\label{fig:NDELS-ablation}}
\end{figure*}

To better understand the interactions of the modules to effectively
dehaze nighttime images, we train each independently and simultaneously.
For a sample image, we evaluate the CLIPIQA , MANIQA, and TRES no-reference,
quality measures \cite{wang_exploring_2022,yang_maniqa_2022,golestaneh_no-reference_2022}.
For all three quality measures, the higher the score the better the
image quality. In Figure \ref{fig:NDELS-ablation}, the DHM appears
to score better than the LLM, as well as combined modules for two
quality measures. However, the TRES quality measure tells us that
the DHM+LLM produces a more pleasing image evidenced by the higher
scores, in particular the alpha-blended image. We can observe that
the DHM creates a partial-vignette effect around the image, while
the combined modules don't.

\begin{table}[H]
\centering
\caption{Ablation study examining the importance of NDELS modules tested on
the NHM dataset.\textcolor{blue}{\protect\label{tab:Ablation-study-examining}}}

\begin{tabular}{c||c|c|c|c|c|c}
\hline 
\multirow{2}{*}{Modules} & \multicolumn{2}{c|}{Base} & \multicolumn{2}{c|}{EMSR} & \multicolumn{2}{c}{Enhancement}\tabularnewline
\cline{2-7}
 & PSNR & SSIM & PSNR & SSIM & PSNR & SSIM\tabularnewline
\hline 
LLM & 13.8821 & 0.7013 & 11.3590 & 0.4731 & 12.9873 & 0.5995\tabularnewline
\hline 
DHM & 14.5164 & \textbf{0.7036} & 11.7811 & 0.4706 & 13.5823 & 0.6024\tabularnewline
\hline 
LLM+DHM & \textbf{14.6458} & 0.7035 & 11.6899 & 0.5078 & 13.4844 & 0.6250\tabularnewline
\hline 
\end{tabular}
\end{table}

Furthermore, upon evaluating each module on the NHM dataset it's evident
that the combination DHM+LLM produces the best result, a PSNR of 14.6458.
In Table \ref{tab:Ablation-study-examining}, we have also broken
down the scores based on whether or not the application of EMSR and
Enhancement are applied after the module. And, as noted earlier, the
EMSR and enhancement don't work well on synthetic images, as the results
indicate.

\subsection*{Runtime Measurement}

The runtime values in Table \ref{tab:Runtimes-of-nighttime-dehazing},
show that NDELS isn't faster than MRPF. However, MRP and MRPF are
both implemented in C++, making the algorithms faster than their non-compiled
versions. Yet, NDELS only using Python, is 3-4 times faster than MRP
and 1.5-2 times faster than OSFD. Network loading time isn't included,
since in a real-world application the model would be initialized once,
then waits for input. It remains to be seen, in future work, how fast
is a compiled NDELS algorithm.

\begin{table}[h]
\centering
\caption{Runtimes of nighttime-dehazing methods on a $800\times600$-sized
color image, averaged over 5 runs.\textcolor{blue}{\protect\label{tab:Runtimes-of-nighttime-dehazing}}}

\begin{tabular}{c||c|c|c|c}
\hline 
Method & MRP\cite{zhang_fast_2017} & MRPF\cite{zhang_fast_2017} & OSFD\cite{zhang_nighttime_2020} & NDELS\tabularnewline
\hline 
Time (s) & 1.460 & 0.241 & 0.623 & 0.397\tabularnewline
\hline 
\end{tabular}
\end{table}

\subsection*{Limitation}

\begin{figure}[h]
\centering
\begin{tabular}{ccc}
\multicolumn{1}{c}{\begin{cellvarwidth}[t]
\centering
{\tiny PSNR: 11.8033, TRES: 74.7440\smallskip{}
}{\tiny\par}

\includegraphics[viewport=0bp 0bp 480bp 360bp,width=2.8cm,height=2cm]{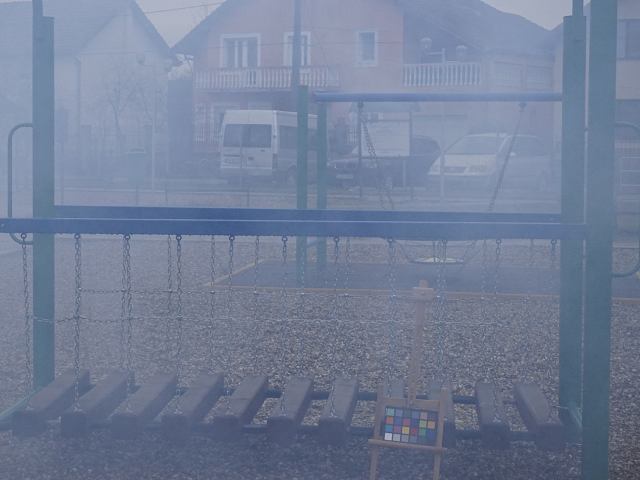}

{\scriptsize (a) hazy}
\end{cellvarwidth}} & \begin{cellvarwidth}[t]
\centering
{\tiny PSNR: 14.5052, TRES: 68.1582\smallskip{}
}{\tiny\par}

\includegraphics[viewport=0bp 0bp 640bp 480bp,width=2.8cm,height=2cm]{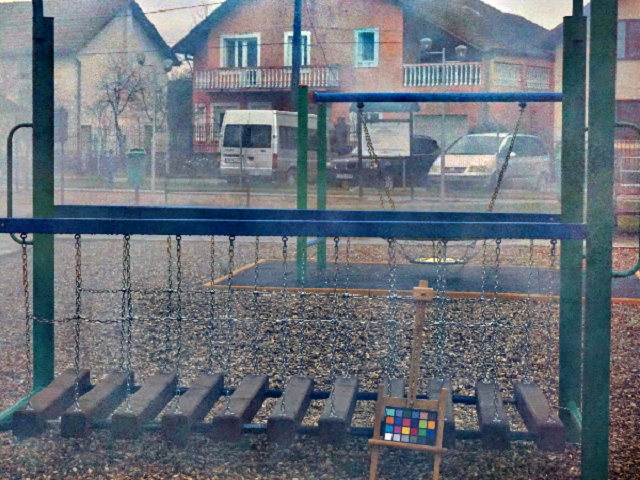}

{\scriptsize (b) MRP\cite{zhang_fast_2017}}
\end{cellvarwidth} & \begin{cellvarwidth}[t]
\centering
{\tiny PSNR: 10.4543, TRES: 67.5944\smallskip{}
}{\tiny\par}

\includegraphics[viewport=0bp 0bp 640bp 480bp,width=2.8cm,height=2cm]{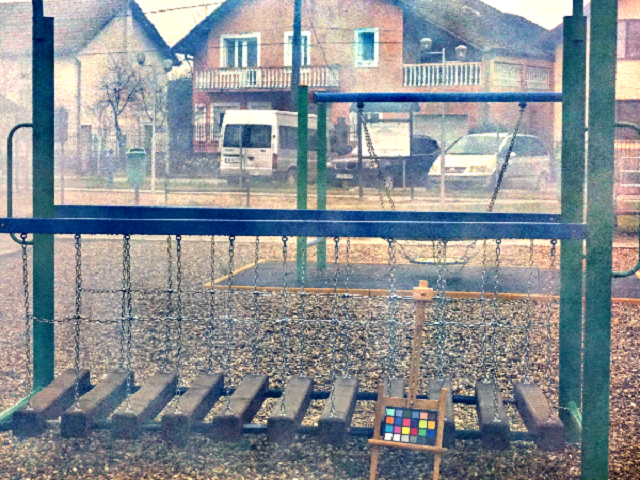}

{\scriptsize (c) MRPF\cite{zhang_fast_2017}}
\end{cellvarwidth}\tabularnewline
\begin{cellvarwidth}[t]
\centering
{\tiny PSNR: 14.8811, TRES: 81.4415\smallskip{}
}{\tiny\par}

\includegraphics[viewport=0bp 0bp 640bp 480bp,width=2.8cm,height=2cm]{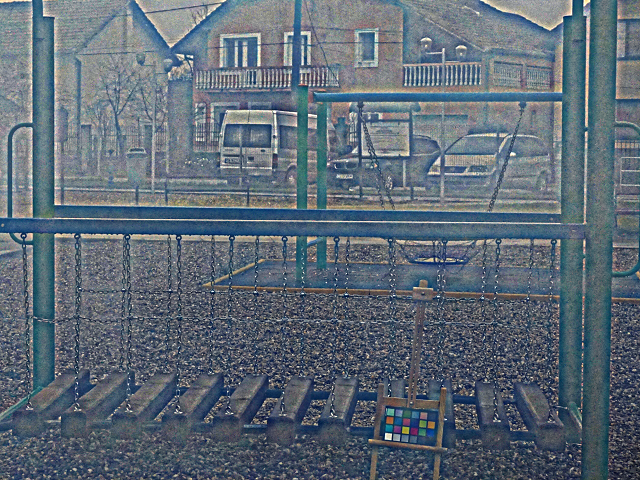}

{\scriptsize (d) OSFD\cite{zhang_nighttime_2020}}
\end{cellvarwidth} & \begin{cellvarwidth}[t]
\centering
{\tiny PSNR: 11.6259, TRES: 81.1322\smallskip{}
}{\tiny\par}

\includegraphics[viewport=0bp 0bp 640bp 480bp,width=2.8cm,height=2cm]{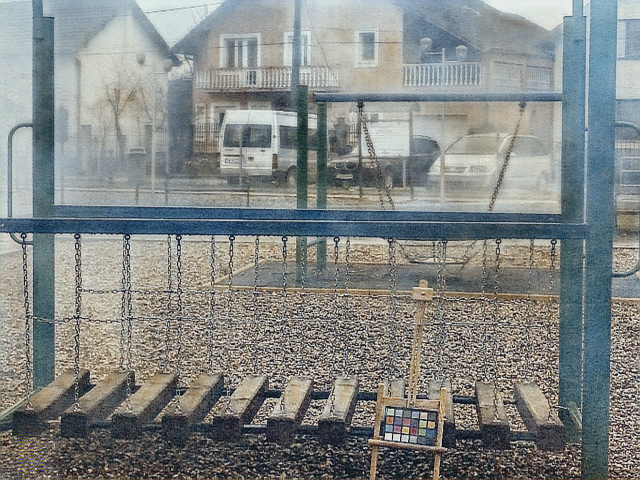}

{\scriptsize (e) NDELS}
\end{cellvarwidth} & \begin{cellvarwidth}[t]
\centering
{\tiny TRES: 95.9874\smallskip{}
}{\tiny\par}

\includegraphics[viewport=0bp 0bp 480bp 360bp,width=2.8cm,height=2cm]{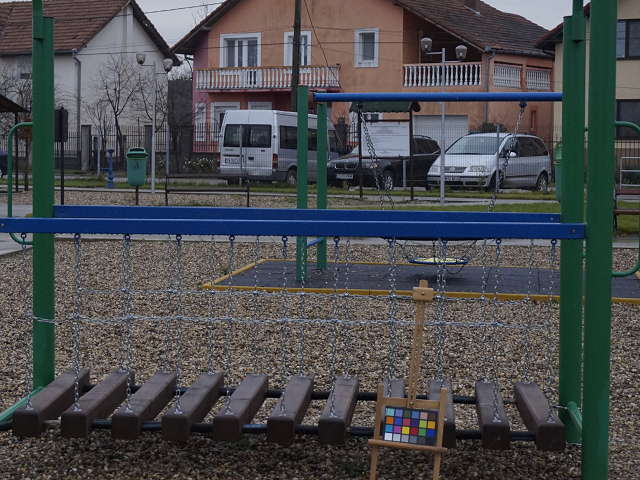}

{\scriptsize (f) GT}
\end{cellvarwidth}\tabularnewline
\end{tabular}

\caption{Nighttime dehazing algorithms tested on a daytime hazy image from
O-HAZE dataset \cite{ancuti_o-haze_2018}.\textcolor{blue}{\protect\label{fig:dehazing-test-of-daytime-images}}}
\end{figure}

During our testing, it became evident that the NDELS network faced
challenges in effectively dehazing daytime images. Notably, this limitation
is not unique to our method, as other nighttime-dehazing approaches
exhibit similar issues, as illustrated in Figure \ref{fig:dehazing-test-of-daytime-images}.
While NDELS does reduce some haze, background objects may appear blurred,
and colors may fade. One potential explanation could be the lack of
training samples with daytime hazy.

For example, in Figure \ref{fig:dehazing-test-of-daytime-images},
OSFD attains the highest PSNR score, but the resulting image appears
dark with bright edges. On the other hand, the TRES scores align more
closely with human-visual perception. As expected, NDELS achieves
higher scores, although counterintuitively, OSFD registers the highest
score. Further investigation is necessary to understand the underlying
factors contributing to this behavior.

\section*{Conclusion}

We present a cutting-edge approach for dehazing and enhancing single-nighttime-hazy
images. The presented method leverages a low-light module (LLM) that
has been trained to suppress light glare and enhance low-light regions
and a dehazing module (DHM) that effectively removes haze, restores
natural color, and sharpens images. Our approach, dubbed NDELS, outperforms
state-of-the-art, single-image nighttime-dehazing, and nighttime-enhancement
methods by better suppressing light effects and enhancing low-light
areas without causing further distortion. Quantitative results on
the NHM dataset show the superiority of NDELS over other approaches
by employing non-reference quality measures to validate the effectiveness
of our extended-multiscale retinex (EMSR) approach. In future work,
we plan to integrate a deraining and desnowing module into the NDELS
network, creating a comprehensive system for nighttime dehazing and
removal of weather elements.

\bibliographystyle{IEEEtran}
\bibliography{paper}

\end{document}